\documentclass[acmsmall]{acmart}
\usepackage{bbding}
\usepackage{ragged2e}
\usepackage{color}
\usepackage{caption}
\usepackage{url}
\usepackage{float}
\usepackage{colortbl}
\usepackage{booktabs}
\usepackage{tablefootnote}
\usepackage{multirow}
\usepackage{makecell}
\usepackage{graphicx}
\usepackage{epstopdf}
\usepackage{float}
\usepackage{verbatim}
\usepackage{array}
\usepackage{tabu}
\usepackage{diagbox}
\usepackage{amsmath}
\usepackage{amsfonts}
\usepackage{geometry}
\usepackage{tabularx}
\usepackage{multirow}
\usepackage{longtable}
\usepackage{adjustbox}
\usepackage{pifont}
\usepackage{makecell}
\usepackage{subcaption}
\usepackage[symbol]{footmisc}

\AtBeginDocument{%
  \providecommand\BibTeX{{%
    Bib\TeX}}}

\renewcommand\footnotetextcopyrightpermission[1]{}
\AtBeginDocument{%
  \providecommand\BibTeX{{%
    \normalfont B\kern-0.5em{\scshape i\kern-0.25em b}\kern-0.8em\TeX}}}
\setcopyright{acmcopyright}
\copyrightyear{2024}
\acmYear{2024}
\acmJournal{JACM}
\acmVolume{V}
\acmNumber{N}
\begin{document}
	

\title{A Survey on Human-Centric LLMs}
% \title{Evaluation and Enhancement of LLMs from a Human-Centric Perspective: A Cross-Disciplinary Survey}

% \title{Bridging AI and Humanities: Evaluating and Enhancing Large Language Models through Human-Centric Perspectives}

\author{Jing Yi Wang*}
\affiliation{
    \institution{Tsinghua University}
    \city{Beijing}
    \country{China}
}
\email{jy-w22@mail.tsinghua.edu.cn}

\author{Nicholas Sukiennik*}
% \authornote{These two authors contributed equally.}
\affiliation{
    \institution{Tsinghua University}
    \city{Beijing}
    \country{China}
}
\email{sukiennikn10@mails.tsinghua.edu.cn}
\thanks{*J. Wang and N. Sukiennik contributed equally to this research.}

\author{Tong Li}
\affiliation{
    \institution{Tsinghua University}
    \city{Beijing}
    \country{China}
}
\email{tongli@mail.tsinghua.edu.cn}

\author{Weikang Su}
\affiliation{
    \institution{Tsinghua University}
    \city{Beijing}
    \country{China}
}

\author{Qianyue Hao}
\affiliation{
    \institution{Tsinghua University}
    \city{Beijing}
    \country{China}
}

\author{Jingbo Xu}
\affiliation{
    \institution{Tsinghua University}
    \city{Beijing}
    \country{China}
}

\author{Zihan Huang}
\affiliation{
    \institution{Tsinghua University}
    \city{Beijing}
    \country{China}
}

\author{Fengli Xu}
\affiliation{
    \institution{Tsinghua University}
    \city{Beijing}
    \country{China}
}

\author{Yong Li}
\affiliation{
    \institution{Tsinghua University}
    \city{Beijing}
    \country{China}
}
\email{liyong07@tsinghua.edu.cn}

\renewcommand{\shortauthors}{Wang et al.}\

\begin{abstract}

The rapid evolution of large language models (LLMs) and their capacity to simulate human cognition and behavior has given rise to LLM-based frameworks and tools that are evaluated and applied based on their ability to perform tasks traditionally performed by humans, namely those involving cognition, decision-making, and social interaction. This survey provides a comprehensive examination of such human-centric LLM capabilities, focusing on their performance in both individual tasks (where an LLM acts as a stand-in for a single human) and collective tasks (where multiple LLMs coordinate to mimic group dynamics). We first evaluate LLM competencies across key areas including reasoning, perception, and social cognition, comparing their abilities to human-like skills. Then, we explore real-world applications of LLMs in human-centric domains such as behavioral science, political science, and sociology, assessing their effectiveness in replicating human behaviors and interactions. Finally, we identify challenges and future research directions, such as improving LLM adaptability, emotional intelligence, and cultural sensitivity, while addressing inherent biases and enhancing frameworks for human-AI collaboration. This survey aims to provide a foundational understanding of LLMs from a human-centric perspective, offering insights into their current capabilities and potential for future development.

\end{abstract}

% \renewcommand{\thefootnote}{\fnsymbol{footnote}}
% \footnote[1]{These two authors contributed equally.}

\maketitle

\section{Introduction}
\label{sec:Introduction}

As large language models (LLMs)~\cite{chang2024survey, zhao2024explainability}, such as OpenAI's GPT family \cite{radford2018improving, radford2019language} and Meta's LLaMA \cite{touvron2023llama, touvron2023llama2}, continue to evolve, their ability to simulate, analyze, and influence human behavior is growing at an unprecedented rate. These models can now process and generate human-like text and perform cognitive tasks at levels comparable to humans in many situations, providing new tools for understanding human cognition, decision-making, and social dynamics.

As such, this survey aims to provide a comprehensive evaluation of LLMs from a human-centric perspective, focusing on their ability to simulate, complement, and enhance human cognition and behavior, both on an individual and collective level. While LLMs have traditionally been rooted in computer science and engineering~\cite{acharya2023llm, kan2024mobile}, their increasing sophistication in replicating human-like reasoning, decision-making, and social interactions has expanded their use into domains where humans are the focal point. This has allowed researchers to address questions that were once too intricate or abstract for computational analysis. For example, in political science, LLMs are used to analyze political discourse, detect biases, and model election outcomes~\cite{rotaru2024artificial}; in sociology, they assist in understanding social media conversations, public sentiment, and group behaviors~\cite{shen2023shaping}; and in psychology, they help model human cognition and decision-making~\cite{demszky2023using}. LLMs have also revolutionized linguistics by enabling large-scale analysis of language, from syntax and semantics to pragmatics~\cite{uchida2024using}, and in economics, they allow for modeling complex interactions between policies and societal outcomes~\cite{li2024econagent}.

To structure this investigation, the survey is divided into two main sections. First, we evaluate human-centric LLMs, focusing on their cognitive, perceptual, social, and cultural competencies. This section examines how LLMs perform tasks commonly associated with human cognition, such as reasoning, perception, emotional awareness, and social understanding. We assess their strengths in structured reasoning, pattern recognition, and creativity, while identifying their limitations in areas such as real-time learning, empathy, and handling complex, multi-step logic. By benchmarking LLM performance against human standards, we highlight areas where LLMs excel and where further improvements are needed.

Second, we explore LLMs in human-centric applied domains, where LLMs are used in real-world scenarios that traditionally require human input. This section is divided into studies focusing on individual and collective applications, where individual-focused studies involve an LLM performing tasks typically done by a single human, such as decision-making, problem-solving, or content creation, and collective-focused studies explore how multiple LLMs can work together to simulate group behaviors, interactions, or collaborative tasks, offering insights into social dynamics, organizational behavior, and multi-agent coordination. In both contexts, we examine the methods employed such as basic prompting, multi-agent prompting, and fine-tuning, along with the theoretical frameworks that guide these applications, including game theory, social learning theory, and theory of mind, etc.

\begin{figure}[tb]
    \centering
    \begin{subfigure}[t]{0.8\textwidth}  % Left figure takes 45% width
    \includegraphics[width=\textwidth]{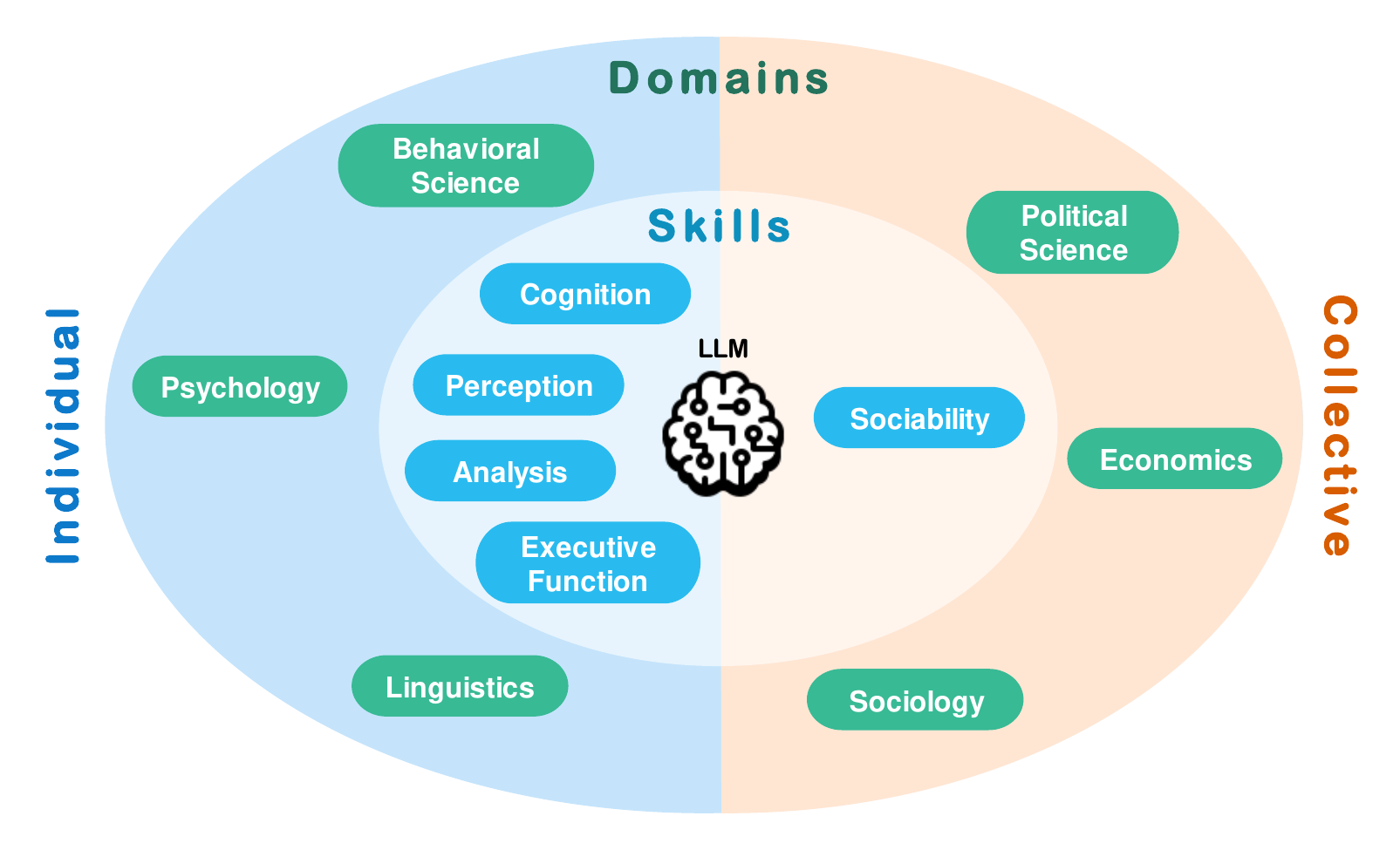}
    
        % \caption{}
        \label{fig:left}
    \end{subfigure}%
    \hfill  % Add horizontal fill between subfigures
    % \begin{subfigure}[t]{0.53\textwidth}  % Right figure takes 55% width
    %     \centering
    %     \includegraphics[width=\textwidth]{Figure/Structure.png}
    %     \caption{}
    %     \label{fig:org}
    % \end{subfigure}
    \caption{ 
    % \textbf{A} 
    Our framework depicts how LLMs are evaluated on foundational human-like skills, divided into individual (e.g., cognition, perception, analysis, executive functioning) and collective (e.g., sociability) levels, and applied within various fields of study similarly categorized as individual (e.g., Behavioral Science, Psychology, Linguistics) and collective (e.g., Political Science, Economics, Sociology) domains.
    % \textbf{B} displays the organization and key topics covered in this survey.
    }
    \label{fig:overview}
\end{figure}

Ultimately, this survey seeks to provide a detailed understanding of how LLMs can better align with human behaviors and social contexts, identifying both their strengths and areas for improvement. Figure~\ref{fig:overview} provides an overview of this framework, categorizing LLM capabilities into individual skills, such as cognition, perception, analysis, and executive functioning, and collective skills like social abilities, and highlighting their capabilities in applying to studies across individual domains like behavioral science, psychology, and linguistics, and collective domains including political science, economics, and sociology. In classifying research works with this framework, we offer insights into how LLMs can be made more effective, ethical, and realistic tools for research and practical applications, whether in individual or collective human-centric settings.

The main contributions of this paper can be summarized as follows. 

\begin{itemize}

\item We provide an in-depth evaluation of LLM capabilities in human-centric tasks, focusing on their cognitive, perceptual, and social competencies, and comparing their performance to human-like reasoning, decision-making, and emotional understanding.

\item We explore LLM's capabilities in human-centric domains, namely focusing on real-world applications in individual and collective contexts, assessing their ability to replicate human behaviors in fields such as behavioral science, political science, economics, and sociology, both as single-agent models and in multi-agent systems.

\item We identify key challenges and future research directions, including improving LLMs' real-world adaptability, emotional intelligence, and cultural sensitivity, while addressing biases and developing more advanced frameworks for human-AI collaboration.

\end{itemize}

The paper is organized as follows: Section 2 provides an overview of AI-empowered human-centric studies and LLMs, while Section 3 evaluates LLM competencies across cognitive, perceptual, analytical, executive, and social skills. Section 4 discusses how LLMs can be applied in a variety of interdisciplinary scenarios to both enhance LLM development and assist in human-centered tasks. Section 5 explores open challenges and outlines future directions for advancing LLMs. Section 6 summarizes key insights and emphasizes the importance of interdisciplinary collaboration to enhance LLMs' understanding of human behavior.

% Finally, Section 5 summarizes the major insights and suggests new directions for future research.

% \textbf{A} depicts a conceptual framework showing the ways in which LLMs interact with human-centric studies and vice-versa. Namely, LLMs serves to help researchers in explaining human phenomena, assist with tasks typically done by humans, and emulate human behavior in scenarios where it can help save humans effort, whereas human studies can then be infused into LLM architectures to make them act more realistically.
% is an overview of LLMs and their interaction with human-centric studies
\section{Overview}
\label{sec:overview}
\subsection{Human-Centric Artifical Intelligence}

\subsubsection{Traditional AI Approaches in Human-Centric Studies}
The application of AI in various human-centered fields has undergone a long progression, now reaching a pinnacle with the rise of generative models, with AI methods to being used investigate various human phenomena. however, despite their relative naivety compared to LLMs, those traditional methods have nonetheless enabled researchers to address complex social phenomena through computation.

% rewrite begins 

 For almost as long as it has been investigated, AI has been used in areas that are highly impactful on society \cite{wooldridge2021brief}. Since then researchers have evaluated the many ways in which AI could emulate human behavior and thought procession, for example in cognition\cite{ullman1978ai}, perception \cite{zadeh2001new}, and executive function \cite{stachowiczstanusch2020management}. More recently, though, with the rise of the web and social media, AI's uses come closer to our day-to-day lives. For example, in political communication research, the detection of political bias in news articles has emerged as a critical area of study, particularly given the increasing polarization in media and online spaces. Traditional methods for predicting political ideology, based on statistical modeling and network analysis, have become an urgent task due to the vast amount of content produced daily. For instance, research by \cite{barbera2015tweeting} employed network analysis to estimate ideological preferences of social media users. Moreover, techniques like topic modeling and content analysis have been widely used to identify bias and misinformation in news articles using data-mining methods \cite{yu2008classifying, shu2017fakea}, highlighting the use of traditional AI techniques in understanding political discourse. Other works tackled the task of stance detection using methods like recursive neural networks \cite{iyyer2014political} and clustering algorithms \cite{darwish2020unsupervised}. Furthermore, Dezfouli \textit{et al.}\cite{dezfouli2020adversarial} explore adversarial vulnerabilities in decision-making models, which is crucial when considering the robustness of traditional bias detection systems under adversarial conditions.
 Furthermore, Dafoe \textit{et al.}\cite{dafoe2020open} emphasize the importance of systems designed to navigate social environments, such as political discourse, using more established multi-agent systems and game theory frameworks. Meanwhile, machine understanding of human preferences has also been used to optimize the learning of reward functions in reinforcement learning\cite{christiano2017deep}, showing us that AI methods not only help us explain human behavior, but can benefit by understanding them, highlighting the co-evolutionary nature of advancements in both AI techniques and human-centric studies. 
% Fine-grained analysis of propaganda in news articles has also been a focus of traditional AI approaches, another critical tool for making online spaces healthy and democratic. Dasanmartino \textit{et al.} \cite{dasanmartino2019finegrained} propose a novel task of detecting propaganda techniques at the fragment level, leveraging feature-based classifiers and neural networks, whereas
% rewrite ends

Overall, the vast body of AI-empowered human-centric studies point to the burgeoning potential of using more advanced computational methods, such as LLMs, to both understand and better simulate human behavior and reasoning processes. LLMs can present new opportunities in the field by simulating human behaviors in areas where real-world data is scarce, as well as facilitate inquiry into laws and dynamics of human behavior based on LLM replicability. 

\subsubsection{A Paradigm Shift from Traditional AI to LLMs}

The rise of LLMs has transformed natural language processing (NLP) and artificial intelligence in general through key breakthroughs in model architecture, scale, and capabilities. Early models like Word2Vec and GloVe used word embeddings, but the introduction of the Transformer in 2017 \cite{vaswani2017attention}, with its self-attention mechanism, enabled deeper contextual understanding and marked a turning point. OpenAI's GPT series, beginning in 2018 with GPT \cite{radford2018improving}, capitalized on this, culminating in GPT-3 \cite{brown2020language} and GPT-4 \cite{achiam2023gpt}, which demonstrated unprecedented capabilities in reasoning, text generation, and multimodal tasks. Meanwhile, Google’s PaLM 2 \cite{anil2023palm} advanced multilingualism and efficiency, and open-source models like Falcon \cite{almazrouei2023falcon} and Baidu’s ERNIE Bot \cite{sun2019ernie} broadened access and specialization. These developments reflect the growing impact of LLMs across diverse domains, from interdisciplinary research to ethical AI applications.

The rapid adoption of LLMs across academic disciplines has led to varying predictions about whether these systems will eventually match human cognitive abilities. While some experts foresee AI achieving human-like general intelligence in the near future, others remain more cautious, doubting whether AI can fully replicate the complex, abstract reasoning and creativity that define human cognition \cite{siemens2022human}. Despite these differing viewpoints, AI is already a significant force in everyday life, influencing decision-making and information processing across numerous domains. However, a key distinction remains: human cognition is driven by forward-thinking, theory-based reasoning, while AI operates on patterns derived from vast datasets, often relying on probability and past data \cite{felin2024theory}. This difference underscores the complementary nature of human and AI systems, with each excelling in distinct aspects of cognitive processing.

Unlike human intelligence, LLMs operate without inherent goals, values, or emotional experiences. Human cognition, driven by survival, social interaction, and creativity, is deeply connected to our physical and social environments. Even embodied AI, while capable of interacting with its surroundings, lacks the nuanced, purpose-driven intelligence that defines human thought. In contrast, LLMs generate responses based on probabilistic models derived from large datasets, without the lived experiences that inform human decision-making. Though LLMs can simulate certain human-like behaviors, they still fall short of the embodied understanding humans possess.

These distinctions raise critical questions about the limitations and potentials of AI, especially as we consider the diverse capabilities explored in Section \ref{sec:LLM_capabilities}, which discusses the capabilities of LLMs including cognitive, perceptual, social, analytical, executive, cultural, moral, and collaborative skills. Section \ref{sec:studies_improve_llm} delves into how interdisciplinary fields, such as  political science, economics, sociology, behavioral science, psychology, and linguistics, contribute to LLM development, offering insights into how human intelligence informs and shapes the evolution of artificial systems. This exploration emphasizes the importance of leveraging LLM strengths while recognizing the fundamental differences between human and artificial cognition.

\section{Evaluation of Human-Centric LLMs}

% \section{Evaluating LLMs Through a Human-Centric Lens}
\label{sec:LLM_capabilities}

% \newcolumntype{L}[1]{>{\raggedright\arraybackslash}p{#1}}
% \newcolumntype{Y}{>{\raggedright\arraybackslash}X} % New column type for text wrapping

\begin{figure}[ht]
    % \vspace*{-10px}
    \centering
    % \hspace{-3mm}
    \includegraphics[width=0.7\linewidth]{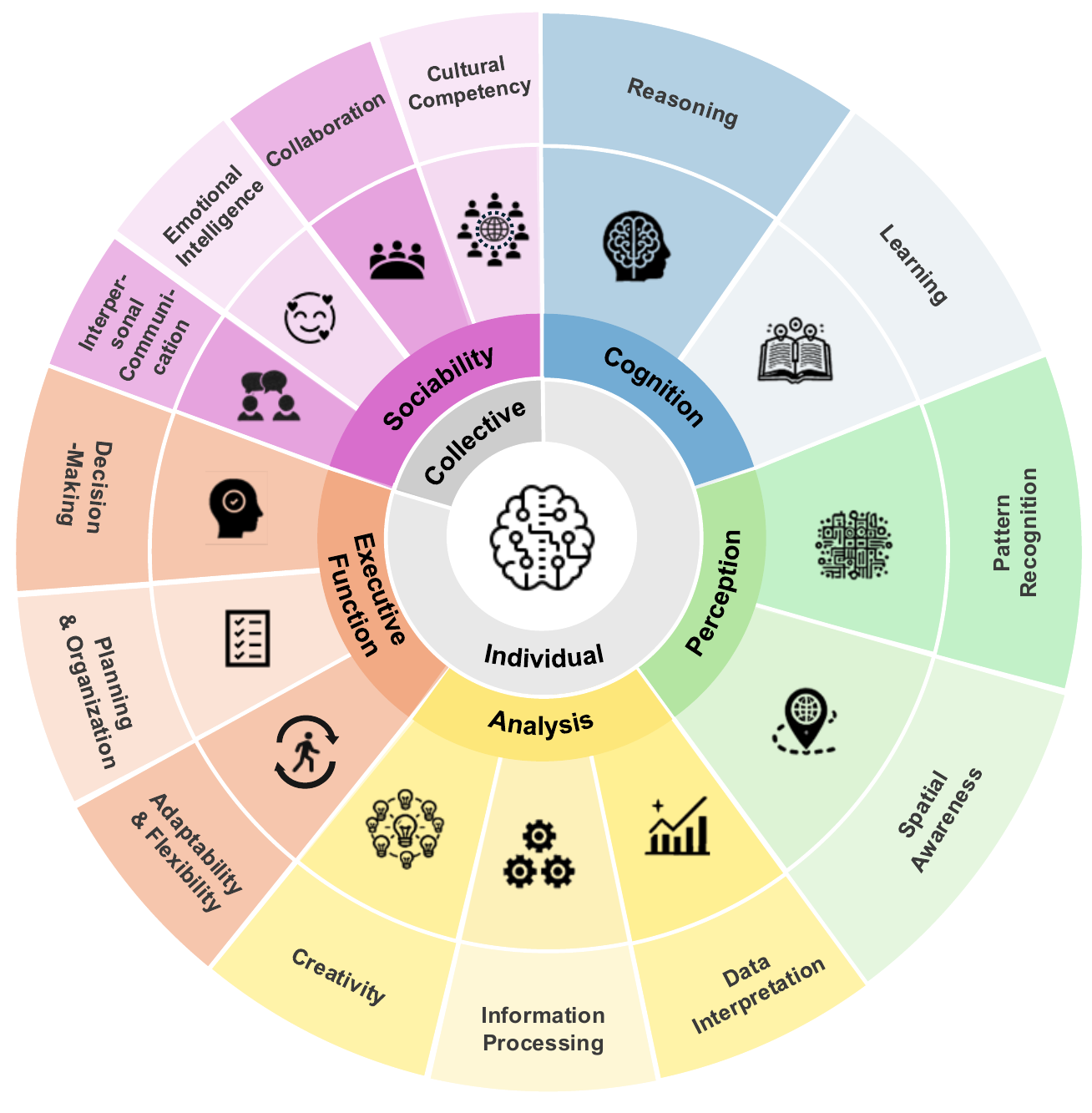}
    % \hspace{-3mm}
    \vspace*{-8px}
    \caption{Overview of LLM Capabilities Across Individual and Collective Domains.}
    \label{fig:llm_competency_diagram}
    \vspace*{-8px} 
\end{figure}

\begin{figure}[!t]
    % \vspace*{-10px}
    \centering
    % \hspace{-3mm}
    \includegraphics[width=0.75\linewidth]{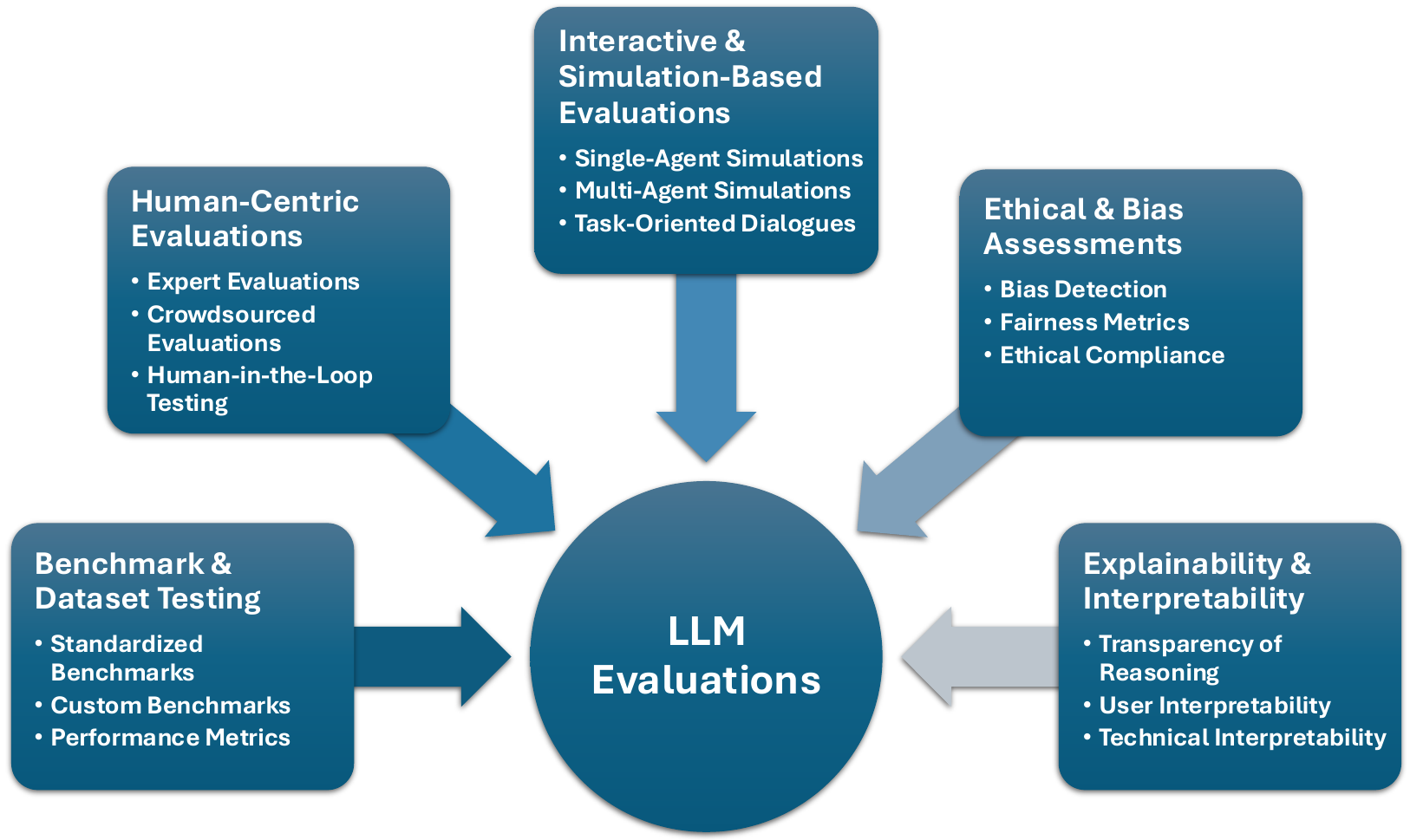}
    % \hspace{-3mm}
    \vspace{-5px}
    \caption{Overview of LLM evaluations. }
    \label{fig:llm_competency_evaluation}
    \vspace{-5px} 
\end{figure}

To evaluate human-centric LLMs, we showcase a holistic representation of LLM competencies, categorized into two domains: individual (e.g., cognitive, perceptual, analytical, executive functioning skills) and collective (e.g., social skills), as shown in Figure \ref{fig:llm_competency_diagram}. This representation includes various key LLM skills, such as reasoning, pattern recognition, spatial awareness, adaptability, decision-making, interpersonal communication, and cultural competency. Following this, Figure \ref{fig:llm_competency_evaluation} outlines the evaluation approaches used to assess LLMs, including benchmark and dataset testing, human-centric evaluations, interactive and simulation-based evaluations, ethical and bias assessments, and lastly, explainability and interpretability evaluations. Table \ref{table:llm_competencies} highlights both the strengths and areas for improvement in these domains. By outlining these abilities, we provide a comprehensive comparison of human-like skills, using benchmarks to assess their strengths and limitations. 
% This structured approach not only reveals where LLMs show  but also pinpoints where they fall short of human cognitive flexibility and adaptability. 
Additionally, Appendix Tables \ref{table:paper_summary} and \ref{table:paper_summary2} provide a comprehensive overview of key papers, highlighting their contributions, the LLMs assessed, and comparisons to human performance. The subsequent section delves into each category, providing an in-depth exploration of the skills and benchmarks that define LLM performance across these domains.

\subsection{Cognitive Skills}
LLMs demonstrate cognitive competencies that mirror key elements of human intelligence, primarily through reasoning and learning. While LLMs show remarkable ability in processing vast amounts of information and generating coherent responses, their proficiency varies when it comes to complex cognitive tasks. These models showcase evolved abilities in structured reasoning and generalization but encounter challenges when faced with intricate logic or learning from real-time interactions. This section explores the strengths and limitations of LLMs in reasoning and learning, highlighting their progress and areas that require further advancement.

% LLMs demonstrate a range of cognitive competencies that mirror key elements of human intelligence, such as reasoning and learning. While LLMs are impressive in their ability to handle vast amounts of data and generate coherent responses, their proficiency in complex cognitive tasks varies. From reasoning to learning, LLMs possess unique strengths but also face limitations in replicating the depth and flexibility of human cognition. In this section, we explore how LLMs perform across these cognitive domains, highlighting both their capabilities and areas for improvement.

\subsubsection{Reasoning}
Logical reasoning, a core element of human cognition and essential for daily functioning, consists of various types of reasoning, including deductive, inductive, and causal reasoning, each contributing to how we process information and make decisions. Deductive reasoning applies general principles to obtain specific conclusions, while inductive reasoning draws generalizations from specific observations~\cite{goswami2010inductive}, and causal reasoning helps to understand cause-and-effect relationships~\cite{eells2016rational, binz2023using}.

Several benchmark datasets have been developed to assess these reasoning capabilities in LLMs. For deductive reasoning, the LogiQA 2.0 dataset~\cite{liu2023logiqa} is a notable resource, focusing on five types of reasoning, including categorical, necessary conditional, sufficient conditional, conjunctive, and disjunctive reasoning. PrOntoQA~\cite{saparov2022language} also evaluates deductive reasoning through first-order logic tasks where LLMs derive specific conclusions from logical premises. For inductive reasoning, CommonsenseQA 2.0~\cite{talmor2022commonsenseqa} requires generalization from everyday facts and commonsense knowledge, whereas the Creak dataset~\cite{onoe2021creak} further tests LLMs’ ability to generalize from commonsense knowledge to identify inconsistencies. In turn, causal reasoning is assessed using CausalBench~\cite{wang2024causalbench}, which evaluates LLMs' ability to reason about cause-and-effect relationships across diverse domains. ContextHub~\cite{hua2024disentangling}, on the other hand, serves as another benchmark focusing on LLMs' causal reasoning in both abstract and contextualized tasks. Additional datasets like GSM8K~\cite{cobbe2021training} and  BIG-Bench-Hard~\cite{suzgun2022challenging} are furthermore employed for mathematical reasoning and evaluating LLM performance across various reasoning domains, respectively. 
%, including multi-step, deductive, inductive, and commonsense reasoning.

Analyzing LLM performance with these datasets has revealed significant insights into their reasoning abilities and limitations. For deductive reasoning, although LLMs like GPT-3 have made progress, their accuracy remains at 68.65\% in tasks involving logical inference, which is significantly below the 90\% human benchmark~\cite{liu2023logiqa}. This gap indicates ongoing challenges in mastering complex logical structures, especially when multiple logical steps or intricate reasoning processes are required. LLMs like GPT-3.5, PaLM, and LLaMA perform well on simpler deductive reasoning tasks but struggle with more complex scenarios that involve chaining multiple logical premises together~\cite{saparov2024testing}. For inductive reasoning, on the other hand, GPT-4 shows improvements in rule application with up to 99.5\% partial accuracy~\cite{bowen2024comprehensive}, yet struggles with larger problems and minimal examples. Even with Chain-of-Thought (CoT) prompting, GPT-4 and Davinci face difficulties in rule validation and integrating complex rules, with Davinci's accuracy declining to 51\% in nuanced tasks~\cite{han2024inductive}. In addition, Han \textit{et al.} \cite{han2024inductive} evaluate GPT-3.5 and GPT-4 on property induction tasks, highlighting that while GPT-4 more closely aligns with human reasoning patterns, they still struggle to fully capture premise non-monotonicity, a critical element of human cognitive processing.

Causal reasoning remains a significant challenge for LLMs like GPT-4 and Davinci, as it requires a deep understanding of cause-and-effect across various contexts. Although these models show reasonable proficiency in mathematical causal tasks, the CausalBench benchmark highlights their struggles with more complex text-based and coding-related causal problems~\cite{wang2024causalbench}. Interpreting causal structures in narratives or code snippets often goes beyond simple data correlations, demanding robust reasoning to avoid producing misleading outputs. Even when GPT-4 initially performs well, its reasoning capabilities frequently weaken when faced with flawed or conflicting arguments, raising concerns about its consistency in complex scenarios~\cite{wang2023can}.

The ContextHub benchmark is developed to assess LLMs like GPT-4, PaLM, and LLaMA in handling both abstract and contextualized logical problems~\cite{hua2024disentangling}. ContextHub focuses on the challenges these models encounter when transitioning from simple logic tasks to nuanced, real-world reasoning. While models perform well with straightforward problems, they often struggle to generalize in context-rich scenarios requiring deeper interpretative skills. Additional datasets like GSM8K emphasize deductive reasoning, and BIG-Bench-Hard evaluates multi-step reasoning, factual knowledge, and commonsense understanding~\cite{cobbe2021training, suzgun2022challenging}. Together, these benchmarks reveal critical insights into the strengths and limitations of models like GPT-4 and Davinci, pinpointing areas that need improvement for handling complex, real-world reasoning tasks.

Overall, these benchmark datasets provide a comprehensive evaluation framework for assessing LLMs' reasoning capabilities, revealing both their advancements and limitations. While LLMs have shown progress in handling specific reasoning tasks, they continue to face significant challenges in multi-step logic, contextual problem-solving, and generalizing their reasoning abilities across diverse domains.

\subsubsection{Learning}

LLMs' learning ability encompasses their capacity to adapt, generalize, and improve performance based on pre-existing training data and interactions with users or environments. Unlike traditional learning models, LLMs do not update their parameters during inference. Instead, they rely on pre-trained knowledge to perform few-shot or zero-shot tasks, highlighting their generalization capabilities. However, this comes with significant limitations when faced with evolving, real-world data.

Recent efforts have aimed at improving LLM adaptability through various strategies. For instance, the RL with Guided Feedback (RLGF) framework \cite{chang2023learning} optimizes learning from feedback, showing that guided strategies can significantly improve text generation in dynamic conditions. Similarly, error-driven learning approaches, like LEMA (Learning from MistAKes) \cite{an2023learning}, allow models like GPT-4 to refine reasoning by identifying and correcting errors. These approaches highlight the potential of leveraging feedback and error correction to boost adaptability, yet they still rely on static data at inference.

While these feedback-based methods provide immediate gains in performance, they do not address the broader challenge of continual learning in real-time. Jovanovic \textit{et al.} \cite{jovanovic2024towards} offer a critical review of incremental learning paradigms, noting that current models, including GPT-4 and GPT-3.5, struggle with dynamic, real-time updates. Though these models excel in handling pre-defined tasks, their performance declines on more complex, evolving instances, reflects a broader limitation of inability to adapt to real-time adaptability.
% , where LLMs cannot update incrementally as new data becomes available.

The fine-tuning approaches explored by Ren \textit{et al.} \cite{ren2024learning}, which aim to clarify instruction-based and preference-based learning, also highlight challenges in maintaining model alignment without overfitting. Although models show improvement through fine-tuning, they still fall short in dynamic environments where adaptability and continual learning are crucial.

In general, these studies reveal both the strengths and limitations of LLMs in learning. While frameworks like RLGF \cite{chang2023learning} and LEMA \cite{an2023learning} demonstrate short-term improvements in adaptability and error correction, and fine-tuning helps refine models, the inability to learn incrementally and in real-time remains a critical obstacle. Future research will need to focus on overcoming this gap to enable LLMs to fully adapt to real-world, evolving tasks, which is why several online-deployed models have begun to include a web search component to collect more recent relevant information \cite{shi2024know, 
xiong2024when}.

\begin{table}[t]
% \newcolumntype{L}[1]{>{\raggedright\arraybackslash}p{#1}}
% \newcolumntype{Y}{>{\raggedright\arraybackslash}X} % New column type for text wrapping
\caption{Summary of LLM Competencies and Benchmarks}
\scriptsize 
\resizebox{\textwidth}{!}{%
\begin{tabular}{p{2.4cm}p{1.8cm}p{7.3cm}p{4.4cm}}
\toprule
\textbf{Category} & \textbf{Skill} & \textbf{LLM Competency} & \textbf{Benchmarks} \\ \midrule

\multirow{4}{*}{\textbf{Cognitive Skills}} 
    & \multirow{1}{*}{Reasoning} 
    & \multirow{1}{=}{LLMs perform well in simpler tasks like structured, rule-based, and abstract logic as models become more advanced, but they struggle with more complex challenges such as multi-step reasoning, handling multiple premises, and adapting to contextual or generalization-based reasoning.}
    & Deductive reasoning: LogiQA 2.0~\cite{liu2023logiqa}, PrOntoQA~\cite{saparov2022language}; \\ 
    % \cmidrule(lr){4-4}
    & & & Inductive reasoning: CommonsenseQA 2.0~\cite{talmor2022commonsenseqa}, Creak~\cite{onoe2021creak}; \\ 
    % \cmidrule(lr){4-4}
    & & & Causal reasoning: CausalBench~\cite{wang2024causalbench}, ContextHub~\cite{hua2024disentangling}; \\ 
    % \cmidrule(lr){4-4}
    & & & Mathematical reasoning: GSM8K~\cite{cobbe2021training}; \\ 
    % \cmidrule(lr){4-4}
    & & & General reasoning: BIG-Bench-Hard~\cite{suzgun2022challenging} \\ 
    \cmidrule(lr){2-4}
    % & Reasoning 
    % & LLMs excel in simpler tasks like structured, rule-based, and abstract logic as models become more advanced, but they struggle with more complex challenges such as multi-step reasoning, handling multiple premises, and adapting to contextual or generalization-based reasoning.
    % & Deductive reasoning: LogiQA 2.0~\cite{liu2023logiqa}, PrOntoQA~\cite{saparov2022language} 
    
    % \n Inductive reasoning: CommonsenseQA 2.0~\cite{talmor2022commonsenseqa}, StrategyQA~\cite{geva2021did}, Creak~\cite{onoe2021creak}
    
    % \n Causal reasoning: CausalBench~\cite{wang2024causalbench}, ContextHub~\cite{hua2024disentangling}
    
    % \n Mathematical reasoning: GSM8K~\cite{cobbe2021training}
    
    % \n General reasoning: BIG-Bench-Hard~\cite{suzgun2022challenging} \\ 
    % \cmidrule(lr){2-4}
    & Learning & LLMs showcase strengths in generalization and feedback-based adaptation but lack traditional online learning and real-time parameter updates.
    &  CommonGen tasks\cite{chang2023learning} , GSM8K\cite{an2023learning}, MATH\cite{an2023learning}, SVAMP\cite{an2023learning}, ASDiv\cite{an2023learning}, CIFAR-100\cite{jovanovic2024towards}, ImageNet100\cite{jovanovic2024towards}\\
    % \\\cmidrule(lr){2-4}
    % & Memory & LLMs excel in context retention and task adaptation but lack persistent memory and real-time recall of past interactions.
    % & Visual Language Navigation (VLN-CE)~\cite{zhang2023memory} \\
    \midrule

\multirow{3}{*}{\textbf{Perceptual Skills}} 
    & Pattern Recognition & LLMs achieve state-of-the-art performance in 3D shape understanding, approaching human-level performance.
    & ScanQA~\cite{hong20233d} \\
    \cmidrule(lr){2-4}
    & Spatial Awareness & LLMs demonstrate strong spatial awareness in structured virtual environments but struggle with adaptability and decision-making in dynamic real-world contexts, revealing limitations in flexible thinking and real-time processing.
    & Red Dead Redemption 2 (RDR2)~\cite{tan2024towards}, MP3D (Matterport3D)~\cite{zhou2023esc} \\
    \midrule

\multirow{3}{*}{\textbf{Analytical Skills}} 
    & Data Interpretation & LLMs are skilled at data extraction, visualization, and structured analysis but remain limited in handling complex, unstructured data, particularly in nuanced entity recognition and relation extraction, compared to human performance.
    & MOF IE~\cite{dagdelen2024structured}, CORD-19 Direction Extraction~\cite{lahav2022search} \\
    
    \cmidrule(lr){2-4}
    & Information Processing & LLMs are proficient in data integration, synthesis, and filtering, outperforming traditional deep neural networks with their natural language understanding, but they still lag behind humans in contextual and domain-specific comprehension.
    & DOC-BENCH~\cite{zou2024docbench}, Cocktail~\cite{dai2024cocktail} \\
    \cmidrule(lr){2-4}
    & Creativity & LLMs showcase advances in rapid, high-quality idea generation but lack human-like intuition, emotional depth, and true creative originality.
    & Purchase Intent Scores~\cite{xu2024jamplate} \\
    \midrule

\multirow{4}{*}{\textbf{Executive Functioning}} 
    &    Adaptability and Flexibility &  LLMs demonstrate increasing adaptability and flexibility in structured tasks, handling real-time changes, but face significant challenges in dynamic, unpredictable environments and remain vulnerable to manipulation and harmful content extraction. & LLM-CI~\cite{shvartzshnaider2024llm}     \\

    % & Contextual Adaptation & LLMs demonstrate improving contextual adaptation, excelling in structured tasks with real-time adjustments, but they still fall short of human flexibility and decision-making in unpredictable environments.
    % & LLM-CI~\cite{shvartzshnaider2024llm} \\
    \cmidrule(lr){2-4}
    & Planning and Organization & LLMs illustrate competency in generating and adapting plans but require external validation for reliability and logical consistency in complex tasks.
    & Embodied Agent Tasks~\cite{song2023llm} \\
    \cmidrule(lr){2-4}
    % & Adaptability & LLMs excel in flexible decision-making and abstract reasoning but remain vulnerable to manipulation and harmful content extraction.
    % & --- \\
    \cmidrule(lr){2-4}
    & Decision-Making & LLMs show considerable competency in decision-making, particularly in risk assessment and managing uncertainty, yet they struggle with cognitive biases inherent in their training data and require more robust mechanisms for responsible use in high-stakes and uncertain situations.
    & $\gamma$-Bench~\cite{huang2024far}, BIASBUSTER ~\cite{echterhoff2024cognitive} \\
    \midrule

\multirow{4}{*}{\textbf{Social Skills}} 
    & Interpersonal Communication & LLMs demonstrate significant progress in simulating human-like social interactions, excelling in structured scenarios and Theory of Mind tasks, but they still struggle with the depth and adaptability required for nuanced, real-world social dynamics.
    & False Belief Task~\cite{strachan2024testing}, Book Genre Typicality~\cite{le2023uncovering} \\
    \cmidrule(lr){2-4}
    & Emotional Intelligence & LLMs outperform human participants in identifying and interpreting emotions.
    & LEAS~\cite{elyoseph2023chatgpt}, EIBENCH~\cite{zhao2024both}, EmoBench~\cite{sabour2024emobench}, SECEU~\cite{wang2023emotional} \\
    \cmidrule(lr){2-4}
    % & Collaboration & LLMs showcase advances in collaboration and teamwork, but they continue to face challenges in matching the adaptability, strategic flexibility, and contextual awareness of human teams, particularly in complex and dynamic situations.
    % & LLM-Arena~\cite{chen2024llmarena} \\

     & \multirow{1}{*}{Collaboration} 
    & LLMs showcase advances in collaboration and teamwork, but they continue to face challenges in matching the adaptability, strategic flexibility, and contextual awareness of human teams, particularly in complex and dynamic situations.
    & AI-to-AI: LLM-Arena~\cite{chen2024llmarena}, RoCoBench\cite{zhang2024towards}; Human-to-AI: GOVSIM \cite{piatti2024cooperate}, Overcooked-AI \cite{zhang2024towards}\\ 
    \cmidrule(lr){2-4}
    & Cultural Competency & LLMs show promise in understanding and simulating cultural behaviors, especially in recalling factual knowledge, but they struggle with nuanced cultural sensitivity, particularly for underrepresented cultures, where humans consistently outperform them in adapting responses based on complex cultural contexts.
    & BLEND~\cite{myung2024blend}, HCD~\cite{bhatt2024extrinsic} \\
\bottomrule
\label{table:llm_competencies}

\end{tabular}%
}
\end{table}

\begin{figure}[!t]
    % \vspace*{-10px}
    \centering
    % \hspace{-3mm}
    \includegraphics[width=1.0\linewidth]{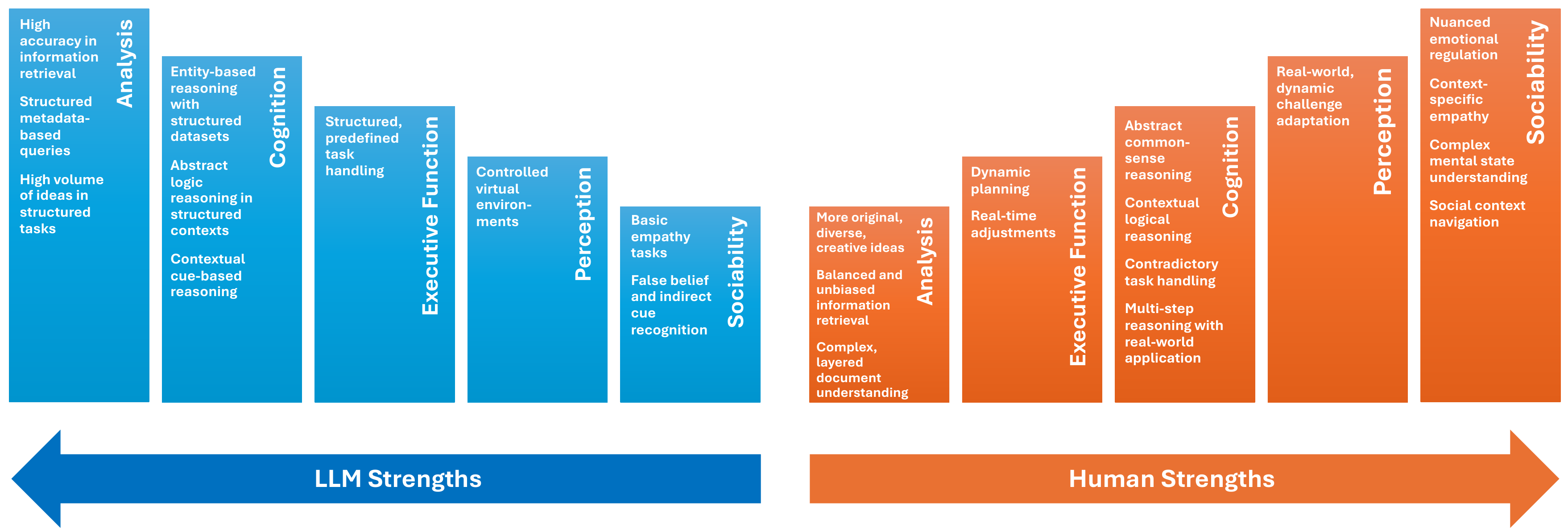}
    % \hspace{-3mm}
    \vspace*{-8px}
    \caption{Competency comparison between LLMs and humans. Blue bars represent skills where LLMs excel over humans, especially in structured tasks and predictable environments, while orange bars indicate skills where humans excel over LLMs, particularly in adaptive, nuanced, and real-world contexts.}
    \label{fig:llm_competency_bar}
    \vspace*{-8px} 
\end{figure}
\subsection{Perceptual Skills}
Perceptual skills in LLMs involve their ability to interpret, organize, and process complex information from their environment, which serves as a foundation for navigating and interacting within both virtual and real-world contexts. These skills are broadly divided into two critical areas: pattern recognition and spatial awareness. Pattern recognition focuses on the LLM's capability to identify and understand recurring structures, semantic relationships, and trends within data, which are essential for tasks involving prediction and interpretation. Spatial awareness, on the other hand, relates to the LLM's ability to comprehend spatial configurations, positions, and relationships between objects within defined environments, crucial for navigation and interaction tasks. Together, these perceptual skills lay the foundation for LLMs to engage effectively in complex scenarios, although challenges remain in achieving human-level proficiency, particularly in dynamic and unpredictable settings.

% Perceptual skills refer to the ability to interpret, organize, and process sensory information from the environment, which are foundational to human cognition. These skills allow humans to make sense of their surroundings, recognize patterns, navigate spaces, and make decisions based on sensory input. Similarly, LLMs exhibit a variety of perceptual skills that enable them to process and interpret complex information, albeit in a way that is grounded in text rather than direct sensory data.
% Existing studies typically evaluate these perceptual skills in LLMs across several domains. First, Environmental Perception examines how well models understand and interpret the broader context or "environment" within text, recognizing relationships, events, or implied information. Next, Pattern Recognition assesses a model's ability to identify and make sense of recurring themes or structures within data, such as linguistic patterns, semantic relationships, or repetitive logic.

\subsubsection{Pattern Recognition}

Pattern recognition refers to a model’s ability to identify, interpret, and categorize recurring patterns within data~\cite{zeng2024similar}. In the context of LLMs, this capability is crucial for recognizing trends, making predictions, and understanding complex multimodal inputs that combine language, vision, and other sensory information. Evaluating pattern recognition involves assessing a model’s proficiency in object recognition, spatial inference, and semantic understanding—core skills required for interpreting both real-world and simulated environments~\cite{zeng2024similar,holzinger2023toward}.

This ability can be analyzed through three critical aspects. Object identification and classification refers to the process where models categorize items based on their distinct features, enabling accurate recognition~\cite{zang2024contextual}. Semantic and spatial relationship mapping involves understanding the relationships between objects and their spatial arrangements, which is essential for contextually accurate reasoning and interaction~\cite{huang2024std}. Finally, temporal dynamics focuses on recognizing sequences of events and their interdependencies, allowing models to grasp patterns over time~\cite{huang2024std}. These aspects highlight the complexity of pattern recognition in LLMs, showcasing their potential while emphasizing the challenges in achieving human-like perception and reasoning.

% This ability can be examined through three key aspects: object identification and classification~\cite{zang2024contextual}, where models categorize items based on features; semantic and spatial relationship mapping, involving the understanding of object relationships and spatial contexts~\cite{huang2024std}; and temporal dynamics, recognizing sequences of events and their dependencies~\cite{huang2024std}. These aspects illustrate the complexity of LLM pattern recognition, highlighting both its capabilities and the challenges in achieving human-like perception.

% Pattern recognition in LLMs can be analyzed from three perspectives:
% 1) Object Identification and Classification: Recognizing and categorizing items based on their features.
% 2) Semantic and Spatial Relationship Mapping: Understanding the relationships between objects, including spatial and contextual cues.
% 3) Temporal Dynamics Understanding: Recognizing sequences of events or actions and understanding their temporal dependencies.

To evaluate these capabilities, several works have employed unique methods
 % large-scale multimodal datasets, including human interaction videos, point clouds, and annotated 3D scene graphs, 
 to help models generalize and recognize patterns in complex environments. For example, the R3M framework by Nair \textit{et al.}\cite{nair2022r3m} employ pre-trained visual representations with time-contrastive learning and video-language alignment, enhancing temporal dynamics in robotic manipulation and achieving over 20\% improvement on benchmarks like Adroit and MetaWorld. Building on these ideas, ConceptGraphs by Qiao \textit{et al.}\cite{gu2024conceptgraphs} integrate 2D foundation model features into 3D scene graphs, advancing tasks like navigation and object manipulation by efficiently recognizing spatial relationships. Additionally, Qi \textit{et al.}\cite{qi2024shapellm} introduce ShapeLLM, which combines visual-language alignment with multi-view image distillation to enhance 3D geometry understanding, useful for embodied interaction tasks like 3D visual grounding. Similarly, 3D-LLM by Hong \textit{et al.}\cite{hong20233d} incorporate 3D spatial understanding, significantly boosting performance in 3D captioning and object navigation using over 1 million 3D-language pairs, whereas Huang \textit{et al.}\cite{huang2022inner} develop the Inner Monologue system, fusing perception and control through real-time feedback to adapt behavior in dynamic environments. Lastly, Voltron by Karamcheti \textit{et al.}\cite{karamcheti2023language} leverage language descriptions and video frames for visual representation learning, surpassing earlier frameworks like MVP and R3M in both low-level pattern recognition and high-level semantic reasoning.

These studies demonstrate the growing integration of visual, spatial, and temporal understanding into LLMs, significantly advancing their ability to recognize complex patterns, generalize across environments, and effectively interact with real-world tasks.

\subsubsection{Spatial Awareness}
Spatial awareness in LLMs refers to their ability to interpret, analyze, and understand spatial configurations, positions, and relationships between objects within defined environments. This skill is essential for tasks involving precise navigation, object identification, and interaction, both in virtual and physical contexts.

LLMs have shown remarkable capabilities in structured virtual environments. For instance, Zhao \textit{et al.} \cite{zhao2023see} develop the STEVE framework within Minecraft, demonstrating LLMs' effectiveness in tasks like block search and tech tree mastery. This highlights LLMs' ability to understand spatial configurations and object relationships in virtual spaces, enabling efficient interaction. Building on these capabilities, Tan \textit{et al.} \cite{tan2024towards} introduce the CRADLE framework, where LLMs interpret visual cues from game screenshots to perform context-aware actions in games like Red Dead Redemption 2. These advancements illustrate significant progress in spatial reasoning, especially in digital environments.

Despite these advances, LLMs face challenges when transitioning from controlled virtual settings to dynamic real-world contexts. Wang \textit{et al.} \cite{wang2023robogen} develop the RoboGen system to evaluate spatial reasoning in physical environments, where LLMs need to adapt to unpredictable scenarios. While they show autonomy in generating robotic tasks and adapting to changes, their performances lack the precision and adaptability of human spatial awareness, especially in real-time adjustments. Liu \textit{et al.} \cite{liu2023aerialvln} explore these challenges further with the AerialVLN framework for UAV navigation in urban environments. Although LLMs effectively follow natural language instructions and recognize spatial landmarks, their handling of complex, real-world navigation remains limited. This emphasizes the gap in LLMs' ability to generalize across diverse and dynamic environments. Schumann \textit{et al.} \cite{schumann2024velma} and Zhang \textit{et al.} \cite{zhang2023building} provide further evidence of these limitations in real-world navigation, showing that while LLMs can interpret spatial cues and adapt to changing conditions, they struggle with continuous decision-making and situational awareness. 
% These findings underscore that although LLMs can recognize spatial relationships, their capacity for dynamic adaptation is underdeveloped compared to human abilities.

The findings from the works about show that LLMs have made important processes in spatial awareness, particularly in structured virtual environments. However, their transition to real-world applications reveals substantial gaps in adaptability and decision-making. They perform well in tasks with predefined rules and clear spatial cues but encounter difficulties in unpredictable environments requiring flexible thinking and real-time processing. Bridging these gaps requires further advancements in contextual adaptability and learning mechanisms in order to bring LLMs closer to human-level spatial proficiency.

\subsection{Analytical Skills}
LLMs demonstrate strong analytical skills in areas such as data interpretation, creativity, and information processing. Below we outline notable works that evaluate LLMs' abilities in each of these areas. 
% data interpretation refers to the model’s ability to analyze and extract meaningful insights from structured and unstructured data, as seen in tasks like data analysis and research synthesis. Creativity entails the model’s capacity to generate novel and unique responses or solutions based on an input, making it useful in tasks such as content creation or problem-solving. Information processing involves efficiently organizing, filtering, and synthesizing large volumes of information, enabling the model to quickly process complex datasets and deliver structured outputs in tasks like report generation or data summarization.

\subsubsection{Data Interpretation}

Data interpretation refers to the ability of LLMs to analyze, interpret, and extract meaningful insights from structured and unstructured data. This capability is crucial in fields such as scientific research, where data-driven decision-making relies on accurately understanding and processing complex datasets \cite{lahav2022search}. Data interpretation in LLMs can be measured through aspects like structured data extraction, data visualization, and reasoning over data to generate actionable insights.

The assessment of data interpretation typically involves evaluating LLMs' ability to handle different data formats, understand domain-specific contexts, and perform accurate computations or visualizations based on the data. Benchmarks such as scientific papers, domain-specific datasets (e.g., chemistry, biology), and tools like MatPlotBench \cite{yang2024matplotagent} are commonly used to assess these abilities, focusing particularly on visualization and data comprehension.

A recent development in this field is the Data Interpreter model by Hong \textit{et al.} \cite{hong2024data}, an LLM-based agent designed to handle complex data science tasks through hierarchical dynamic planning and real-time data adaptation. The model demonstrates significant improvements in managing data dependencies and optimizing machine learning workflows, achieving a 26\% enhancement in mathematical problem-solving and a 112\% improvement in open-ended tasks. These results highlight the possible enhancement for LLMs to adapt to dynamic data relationships, a key challenge in data interpretation.

% In the domain of chemical research, Coscientist, based on GPT-4, was developed to autonomously interpret data from complex experiments \cite{boiko2023autonomous}. Coscientist effectively combined LLM-driven data interpretation with experimental automation to perform tasks such as palladium-catalyzed cross-coupling reactions, demonstrating its ability to optimize experimental conditions and accelerate scientific discoveries. This showcases how LLMs can influence experimental workflows by deriving actionable insights from complex data.

% For data extraction in materials science, ChatExtract by Polak \textit{et al.} \cite{polak2024extracting} represents a notable advancement. ChatExtract utilizes conversational LLMs to extract precise data, such as material properties, from unstructured research papers. Through advanced prompt engineering, it achieves over 90\% precision and recall rates, emphasizing LLMs’ potential to automate the extraction and structuring of scientific information from unstructured sources.

Data visualization, another aspect of data interpretation, has also been addressed using LLMs. MatPlotAgent \cite{yang2024matplotagent} automates the generation of complex visualizations from user queries and raw data, iteratively refining output through feedback. This framework demonstrates the potential of LLMs to interpret data and produce high-quality visual representations, further aiding in data comprehension and analysis.

Dagdelen \textit{et al.} \cite{dagdelen2024structured} explore structured information extraction using LLMs to create knowledge bases from scientific texts. By combining named entity recognition with relation extraction, they successfully transform unstructured research papers into structured data that can be used in downstream applications like machine learning. This contributes significantly to automating the research process, making vast amounts of scientific literature accessible in structured formats.

From the above findings, it is clear that data interpretation is a key capability of LLMs, empowering them to process, analyze, and visualize complex datasets across domains. The assessments conducted in these works highlight LLMs' growing proficiency in extracting and leveraging insights from data, proving their value as helpful assistants to humans in various aspects of data-driven scientific research.

\subsubsection{Information Processing}
LLMs demonstrate advanced analytical capabilities in information processing, distinguishing themselves through their ability to handle data integration, synthesis, and filtering. Unlike previously discussed perceptual skills, which primarily involve interpreting sensory inputs, these analytical functions rely on deeper cognitive processing, leveraging natural language understanding and generation to manage data in sophisticated ways beyond traditional neural networks.

While deep neural networks have proven effective in analytics and management, they often struggle with the generalization and semantic understanding required as data scales up. LLMs overcome these limitations by better processing language, allowing users to communicate with the system in a way that feels natural and efficient. For example, Zhang \textit{et al.} \cite{zhang2023large} demonstrate how LLMs can perform data interpolation and detection through sophisticated prompt techniques, showing how they support human decision-making by simplifying data preprocessing tasks that would otherwise require more manual input. Additionally, interactive systems like NL2Rigel~\cite{huang2023interactive} allow users to refine tables from semi-structured text using natural language instructions, exemplifying how LLMs facilitate intuitive and efficient data synthesis.

LLMs also address challenges such as data scarcity by generating high-quality synthetic data. Auto-regressive generative LLMs, for instance, can create realistic tabular datasets, providing valuable solutions when real-world data is incomplete or unavailable \cite{borisov2022language}. This capability to simulate realistic scenarios allows users to work with more robust datasets, further enhancing decision-making processes.

% practical benefits, especially in environments where data completeness is critical, 
To further evaluate LLM-based document processing, the DOCBENCH benchmark \cite{zou2024docbench} is introduced, specifically designed for assessing LLMs in tasks involving document reading, metadata extraction, and multi-modal information understanding. This benchmark reveals that while LLMs have made significant progress in document processing, they still lag behind human performance, particularly in scenarios that require nuanced comprehension and adaptability across diverse real-world contexts. In information filtering and relevance determination, on the other hand, LLMs surpass traditional recommender systems and search engines by enabling active user engagement and personalized information filtering \cite{galitsky2024llm}. Eigner \textit{et al.} \cite{eigner2024determinants} demonstrate that LLMs enhance decision-making quality in human-AI collaborations by effectively prioritizing and determining relevance in complex data streams.

% Regarding information filtering and relevance determination, LLMs outperform conventional recommender systems and search engines, which typically act as passive information filters. Unlike these traditional systems, LLMs can establish the groundwork for active user engagement \cite{galitsky2024llm}, offering more dynamic and personalized filtering of information based on user inputs. Eigner \textit{et al.} \cite{eigner2024determinants} conducted a comprehensive analysis on decision-making with LLMs, demonstrating how LLMs, supported by a dependency framework, enhance the quality of decisions in human-AI collaborations. By empowering both users and organizations, LLMs contribute to more effective decision-making processes, adding value through their ability to prioritize and determine relevance in complex information streams.

In mixed-source environments, benchmarks such as Cocktail~\cite{dai2024cocktail} assess LLMs' ability to process both human- and machine-generated content, highlighting the complexities of ensuring accuracy and fairness in information retrieval. While LLMs demonstrate strong capabilities in managing large and diverse datasets, they also emphasize the need for a balanced approach to address the variability and biases inherent in mixed-source data.
% This ongoing challenge underscores the importance of human oversight and the nuanced role LLMs play in supporting, rather than fully replacing, human judgment in data-intensive tasks.

In sum, LLMs represent a significant step forward in information processing, offering enhanced capabilities in integrating, synthesizing, and filtering data. Their ability to engage with users in natural, intuitive ways emphasizes a more human-centered approach to technology, one that extends traditional data models by aligning more closely with how humans interact with and make sense of complex information. Yet, as these systems evolve, their ongoing development will need to focus on refining the balance between machine efficiency and human-like adaptability in increasingly dynamic and nuanced environments.

\subsubsection{Creativity}
Large, pre-trained models exhibit remarkable creativity, generating fresh ideas and unique outputs across various domains from art to science, fundamentally transforming the creative industries and pushing the boundaries of human imagination.
Girotra \textit{et al.}\cite{girotra2023ideas} find that LLMs excel in generating innovative ideas, surpassing humans in both speed and quality, even outperforming students from elite universities in creating high-purchase-intent ideas. Similarly, Xu \textit{et al.}\cite{xu2024jamplate} propose an LLM-enhanced design template that promotes critical thinking and idea iteration during the creative process, offering a template to guide users to analyze problems deeply and improve their creativity.
% through a structured format that helps organize and develop ideas.

% Building on the expanding role of LLMs in creativity, 
 % review 110 , emphasizing the growing role of LLMs in creativity and human-computer interaction. LLMs 
After reviewing 110 studies, Li \textit{et al.}\cite{li2024map} discover that LLMs are increasingly used for advanced tasks like software development and creative content generation, lowering barriers to human-AI collaboration, particularly in industries like scriptwriting and advertising. In a related effort to enhance LLM creativity, Lu \textit{et al.}\cite{lu2024llm} propose the ``LLM Discussion'' framework, which enhances creativity through multi-round role-playing discussions. This three-stage process—initiation, discussion, and convergence—guides LLMs to generate more innovative solutions, outperforming both single LLM methods and existing multi-model frameworks in creativity benchmarks. Together, these studies highlight LLMs' transformative potential in innovation and creativity.

Overall, these four studies demonstrate the transformative potential of LLMs in innovation and creativity, showing their superiority in idea generation, enhancing critical thinking, advancing human-AI collaboration, and improving creative output in traditionally human-dominated tasks. 
% through frameworks like design templates and multi-round discussions.

\subsection{Executive Functioning}

Executive functioning in LLMs involves skills like adaptation, planning, and decision-making, enabling models to manage complex tasks and respond to changing conditions. 
% While LLMs show progress in structured tasks, they still struggle with unstructured scenarios requiring human-like flexibility and nuanced decision-making~\cite{luu2024context,kim2023dynacon,shvartzshnaider2024llm}. 
This section examines these aspects of executive functioning, highlighting both strengths and areas for improvement.

\subsubsection{Adaptability and Flexibility}
Adaptability and flexibility are essential components of executive functioning in LLMs, reflecting their ability to adjust actions and strategies as task demands change. 
% While LLMs have shown potential in structured environments, they struggle in unstructured or unpredictable scenarios, where human-like flexibility is crucial.

Contextual Adaptation refers to how well LLMs modify their behavior in response to evolving contexts. Luu \textit{et al.}\cite{luu2024context} demonstrate that LLMs can leverage past experiences to break tasks into context-sensitive components, achieving promising adaptability in controlled settings. However, these models falter in more dynamic, unstructured environments. Similarly, Kim \textit{et al.}\cite{kim2023dynacon} show that LLMs guiding robotic agents through changing environments in the Dynacon system, adjusting to real-time inputs without explicit instructions. Despite this progress, LLMs still struggle to match the nuanced, context-aware decision-making humans exhibit when fast, flexible responses are needed.

To benchmark these capabilities, researchers have employed frameworks like LLM-CI ~\cite{shvartzshnaider2024llm}, which evaluates how well LLMs align with societal norms and navigate human interactions. Although LLMs can partially understand the rules guiding human behavior, they lack depth in contextual understanding, particularly in social settings where subtle cues and ethical considerations are involved. This gap underscores LLMs' broader limitations in adapting to complex, real-world scenarios.

Adaptability also involves handling uncertainty and overcoming obstacles in decision-making. Eigner \textit{et al.}\cite{eigner2024determinants} evaluate LLMs across factors like task difficulty and psychological influences, finding that models perform well in structured contexts but struggle with complexity and unpredictability. Zhang \textit{et al.}\cite{zhang2024large} expose weaknesses by showing how interrogation techniques can force LLMs to produce harmful content even after alignment training, highlighting limitations in adapting to malicious manipulation. On the technical side, the Ctrl-G framework ~\cite{zhang2024adaptable} improves adaptability by combining LLMs with Hidden Markov Models to generate constrained text output. This enhances LLM performance in specific tasks, though their ability to adapt to more open-ended scenarios remains limited.

While LLMs demonstrate increasing adaptability in structured tasks, significant challenges remain in replicating human-level flexibility in real-world situations. Benchmarks like those developed by Luu \textit{et al.} and Kim \textit{et al.} show that LLMs can adjust well in predictable, structured environments, but their limitations become clear in dynamic, uncertain contexts. These findings indicate that while LLMs show potential, their adaptability is still far from human capabilities, particularly in managing complexity and ambiguity. Future research should focus on improving learning mechanisms and algorithms that allow LLMs to better handle uncertainty and mimic human cognitive flexibility.

\subsubsection{Planning and Organization}
LLMs possess advanced planning and organizational capabilities, allowing them to break down complex tasks into smaller, manageable steps, arrange information logically, and generate coherent plans, fostering efficient decision-making and execution. In the context of LLMs, planning and organization refer to different cognitive-like processes that the model can simulate when generating text or solving complex tasks. In particular, planning refers to the model's ability to structure its outputs in a coherent and logical way over multiple steps or turns. Organization refers to how well the model arranges and structures information within a single response or across multiple interactions.

Song \textit{et al.}~\cite{song2023llm} introduce the LLM Planner, a method that uses LLMs to empower intelligent agents to follow natural language instructions and complete complex tasks in visually perceived environments. By learning from a small number of samples, LLM Planner significantly reduces the amount of annotated data needed to train embodied agents. The LLM Planner excels in breaking down complex tasks into smaller, actionable steps for the agent to execute while adapting to real-world challenges. In contrast, Kambhampati \textit{et al.}\cite{kambhampati2024llms} argue that while LLMs are strong in language comprehension and generation, they are not fully capable of effective planning or self-validation. To address these limitations, they propose the LLM Modulo framework, a bidirectional interaction system combining LLMs with external model-based validators. Their roles include generating candidate plans, refining problem specifications, and translating domain models. This framework positions LLMs as incomplete planners that contribute to the broader planning process but depend on external validation for accuracy and reliability. 

Gundawar \textit{et al.} \cite{gundawar2024robust} apply the LLM Modulo framework to travel planning, showing how LLMs can generate itineraries from natural language queries and use external critics for feedback. Their results demonstrated that GPT-4-Turbo improves planning success rates, outperforming traditional methods. Similarly, Sharan \textit{et al.} \cite{sharan2023llm} introduce LLM-ASSIST, which combines rule-based planners with LLMs for autonomous driving. The system leverages LLMs' reasoning to generate safe driving plans for complex scenarios beyond the scope of rule-based systems.

Overall, these studies emphasize the role of LLMs in planning, focusing on dynamic adaptability as well as the importance of external validation to ensure plan feasibility.

\subsubsection{Decision-Making}
In the domain of decision-making, LLMs have both strengths and inherent limitations, particularly in risk assessment, managing uncertainty, and addressing cognitive biases. As LLMs become more widely deployed, their role in supporting decision-making processes must be rigorously evaluated.

In risk assessment, LLMs are expected to accurately identify and manage the risks tied to their integration. Given their growing use in various sectors, vulnerabilities must be carefully addressed. Pankajakshan \textit{et al.}\cite{pankajakshan2024mapping} develop a systematic framework using scenario analysis and dependency mapping to prioritize risks in LLM integration. GUARD-D-LLM \cite{vishwakarma2024guard} further assesses threats tied to specific use cases, highlighting the need for ongoing improvements in risk mitigation frameworks.

% In risk assessment and management, LLMs are expected to accurately assess and manage the risks associated with their integration and use. Given the increasing reliance on LLMs in various sectors, there is a heightened need to address vulnerabilities that can arise from their deployment. Pankajakshan \textit{et al.}\cite{pankajakshan2024mapping} tackled this by combining scenario analysis, dependency mapping, and impact analysis to develop a systematic framework for identifying and prioritizing risks specific to LLM integration. Their work emphasizes the need for LLMs to handle risk in a structured manner, particularly in environments where the consequences of poor decision-making are severe. Complementing this, GUARD-D-LLM \cite{vishwakarma2024guard} was developed to pinpoint and rank threats tied to particular use cases, using text-based user inputs to assess risks. These frameworks underline LLMs' potential in structured risk assessment but also highlight the importance of continuous improvements in the models to ensure reliable risk mitigation.

Evaluating LLMs' decision-making through the lens of game theory, Huang \textit{et al.} \cite{huang2024far} introduce GAMA-Bench ($\gamma$-Bench), a framework that assesses LLMs' decision-making abilities in multi-agent environments using classic game theory scenarios like the Public Goods Game and Sealed-Bid Auction. This benchmark evaluates LLMs on cooperative, betraying, and sequential decision-making, highlighting strengths in robustness but revealing limitations in generalizability compared to human decision-makers. While LLMs like GPT-3.5 show promising robustness, their performance in complex, multi-agent settings still lags behind humans, suggesting the need for approaches like CoT to improve their strategic capabilities.

When it comes to decision-making under uncertainty, LLMs are increasingly used by decision-makers who may not fully grasp the nuances of these complex models. As LLM capabilities evolve, there is a growing temptation to rely on them for solving ambiguous and uncertain problems. However, this introduces risks, including the awareness and potential misuse of LLMs and the systemic risks of relying on AI for critical decisions calling for a dynamic, adaptive approach in AI development, especially when LLMs are used in high-stakes scenarios \cite{pilditch2024reasoning}. 

% It emphasizes the importance of improving technical aspects and implementing robust policies for safe and responsible use, especially in unpredictable environments.

A key challenge in LLM-based decision-making is the presence of cognitive biases. Since LLMs are trained on large datasets that often contain societal biases, they can generate biased responses \cite{zhao2018gender}. Similarly, Schramowski \textit{et al.} \cite{schramowski2022large} note that these biases resemble the cognitive biases seen in human behavior. Such biases can distort decisions and undermine fairness and objectivity. To address this issue, Echterhoff \textit{et al.} propose a self-debiasing technique that enables LLMs to automatically adjust prompts, removing bias-inducing elements and enhancing the fairness and consistency of their decision-making processes \cite{echterhoff2024cognitive}.
% , marking an important step toward more ethical AI use in decision-making.

Overall, while LLMs show potential in risk assessment, uncertainty management, and decision-making, they still require significant improvements to handle complex scenarios, cognitive biases, and high-stakes environments effectively in order to assist humans effectively.

% Overall, LLMs possess considerable competency in decision-making processes, particularly in risk assessment, managing uncertainty, and improving decision consistency. However, these competencies are tempered by limitations, especially when it comes to dealing with cognitive biases and ensuring that LLMs are used responsibly in uncertain environments.

\subsection{Social Skills}
Social skills in LLMs are essential for enabling effective human-like interactions, emotional understanding, collaboration, and cultural awareness. Key areas such as interpersonal communication, emotional intelligence, collaboration, and cultural competency define how well LLMs navigate social dynamics. The works described below give us an overview of the strengths and limitations of LLMs is such scenarios.

% While there have been significant advancements, LLMs still encounter challenges in handling complex, dynamic, and culturally nuanced situations.

\subsubsection{Interpersonal Communication}

Interpersonal communication in LLMs refers to their ability to engage in human-like social behaviors, which is crucial for effective information exchange, understanding social dynamics, and participating in complex interactions \cite{williams2023epidemic, mahowald2024dissociating, le2023uncovering}. 
% Researchers measure these abilities through various tasks, such as role-playing, simulating group behaviors, and participating in dialogue systems that emulate virtual environments and human social simulations.

To comprehensively assess LLMs' interpersonal communication abilities, the focus is often on two primary dimensions: communication skills and social reasoning. Communication skills are assessed by examining the LLM's ability to produce coherent and contextually relevant responses, while also understanding language subtleties like tone, humor, and implied meanings~\cite{wu2024benchmarking}. Social reasoning involves the model's ability to interpret and predict the thoughts, intentions, and emotions of others~\cite{gandhi2024understanding}. A critical aspect of social reasoning is Theory of Mind (ToM)—the capacity to recognize that others may hold beliefs, desires, or knowledge different from one’s own~\cite{strachan2024testing}. This capability is essential for predicting behavior and responding appropriately in dynamic social interactions.

Various experimental setups and datasets have been developed to test these aspects of interpersonal communication in LLMs. For instance, the Generative Agent Environment \cite{park2023generative} provides a controlled virtual setting where LLM-based agents simulate everyday social behaviors, like forming relationships and making collective decisions. These agents operate in environments modeled after games like The Sims, demonstrating their ability to plan activities, organize events, and engage in realistic social dynamics, which showcases their evolving communication and coordination capabilities. 

Role-playing has emerged as another significant method for assessing LLMs' interpersonal skills. Shanahan \textit{et al.} \cite{shanahan2023role} investigate how LLMs can take on different personas to simulate complex social scenarios, including deception, persuasion, and conflict resolution. This approach highlights the potential of LLMs to be utilized in training environments where realistic social interactions are required, such as in negotiation exercises or interview preparations, further proving their relevance in human-like simulations. Furthermore, Strachan \textit{et al.} \cite{strachan2024testing} explore LLMs' social reasoning capabilities by focusing on ToM tasks. Their research demonstrates that models like GPT-4 could perform well in understanding indirect requests and predicting the beliefs and emotions of others. However, these models still face challenges when dealing with more nuanced social situations, such as identifying faux pas or subtle social cues, indicating areas for further improvement.

Overall, the interpersonal communication capabilities of LLMs are evolving rapidly, with models demonstrating improved proficiency in simulating human-like social behaviors across various domains. While challenges remain in achieving the depth and adaptability of human interactions, especially in nuanced or unpredictable contexts, LLMs' growing competencies make them valuable assets for applications ranging from virtual environments to training simulations and social behavior studies.

% Through tasks like structured role-playing, communication exercises, and ToM evaluations, LLMs are becoming increasingly adept at navigating complex social dynamics， al 

\subsubsection{Emotional Intelligence}
Emotional intelligence (EI) in LLMs is a crucial aspect of their social skills, enabling them to recognize, understand, and respond to users' emotions. In the field of "machine psychology," which integrates human psychology principles into AI systems~\cite{hagendorff2023machine}, EI enhances human-computer interactions by fostering empathetic communication and more engaging responses.

EI in LLMs includes emotion perception, cognition, and expression, all vital for interpreting emotional cues and navigating social contexts. Researchers use psychometric tools like EIBENCH~\cite{zhao2024both} and EmoBench~\cite{sabour2024emobench} to evaluate these abilities. EIBENCH focuses on emotion recognition, causal reasoning, and generating empathetic responses, while EmoBench tests emotional understanding and reasoning in complex social scenarios.

Zhao \textit{et al.}\cite{zhao2024both} develop the Modular Emotional Intelligence enhancement method (MoEI), improving models' EI without reducing their general intelligence (GI). Experiments show that models like Flan-T5 and LLaMA-2-Chat effectively balanced EI and GI, displaying enhanced empathy and emotional reasoning. Wang \textit{et al.}\cite{wang2023emotional} introduce the SECEU test, revealing that models like GPT-4 can outperform human participants in emotion understanding. Elyoseph \textit{et al.}~\cite{elyoseph2023chatgpt} use the Levels of Emotional Awareness Scale (LEAS) to find that ChatGPT exceeds average human performance in emotional awareness, showing improvements through iterative learning. Despite advancements in emotion recognition and regulation, LLMs still lag behind human-level social reasoning in complex scenarios. Frameworks like EmoBench \cite{sabour2024emobench} highlight that while models like GPT-4 perform well in emotional tasks, gaps remain in their ability to handle nuanced social communication.

OVerall, the above works imply that while LLMs demonstrate significant progress in emotional intelligence, especially in perception and empathy, they struggle to fully apply this understanding in real-world social interactions.

% Tools like EIBENCH and EmoBench offer valuable insights into these strengths and limitations, guiding future improvements toward more nuanced social capabilities.

\subsubsection{Collaboration}

LLM collaboration capabilities span two critical domains: working alongside AI agents and collaborating with human partners. These areas represent different aspects of how LLMs contribute to teamwork, strategic planning, and task execution, yet each has distinct challenges and benchmarks for evaluation.

In multi-agent systems (MAS), LLMs like GPT-3.5, GPT-4, and Claude-3 have demonstrated improvements in task execution and strategic decision-making~\cite{liu2023dynamic}. Frameworks such as the Dynamic LLM-Agent Network (DyLAN) allow multiple LLM agents to share information and coordinate strategies, improving reasoning and code generation tasks by up to 13\%~\cite{liu2023dynamic}. However, they also show that while GPT-4 exhibits greater adaptability in strategy formation, models like Claude-3 tend to struggle more with maintaining complex collaborative dynamics, especially in scenarios requiring continuous adjustments.

Zhang \textit{et al.}~\cite{zhang2024towards} introduce the Reinforced Advantage (ReAd) framework, enhancing collaboration in simulated environments through a feedback loop that refines actions based on learned advantage functions. Though ReAd improves strategic planning efficiency, LLMs still exhibit rigidity when faced with unpredictable and dynamic conditions, where adaptive collaboration is crucial.

Meanwhile, benchmarks such as LLMARENA~\cite{chen2024llmarena}, which assesses LLM collaboration in multi-agent environments through TrueSkill™ scoring, illustrate that despite models like GPT-4 showing improvements in structured environments, they struggle to match the real-time adaptability and teamwork exhibited by human agents, particularly in the areas of opponent modeling and dynamic communication.

In human collaboration, LLMs like GPT-4 and LLaMA2-70B contribute effectively to decision-making, resource management, and strategic planning, but they often fail to grasp nuanced social cues and contextual awareness, both of which are essential for seamless human interaction. On platforms like GOVSIM~\cite{piatti2024cooperate}, LLMs typically lag behind human teams in maintaining strategic adaptability, especially in situations requiring quick decision-making and responses to unforeseen developments.

The RoCoBench benchmark~\cite{zhang2024towards}, which evaluates LLM-human collaboration, highlights the limitations in LLMs’ strategic reasoning and interaction efficiency. While frameworks like ReAd provide incremental improvements, RoCoBench results indicate that LLMs remain rigid and task-focused compared to the flexible and intuitive collaboration seen in human teams.

Overall, despite the advancements, LLMs face significant limitations in both AI-agent and human collaboration. In AI-agent interactions, LLMs lack the contextual understanding and flexibility that are vital in dynamic environments, where human reasoning excels. In human collaboration, LLMs struggle to handle the complexity of nuanced social cues and real-time adjustments essential for effective teamwork. Therefore, future advancements should focus on enhancing LLMs' ability to process real-time feedback and evolve strategies dynamically.

\subsubsection{Cultural Competency}
LLMs have shown significant potential in understanding and simulating cultural behaviors, offering promising applications in diverse contexts. However, they face notable challenges in accurately applying this knowledge across cultures, particularly for minority or underrepresented groups. These challenges often lead to biased or stereotypical outputs, undermining cultural sensitivity and inclusivity.

For example, LLMs like GPT-4 Turbo have been tested for generating culturally relevant content in languages such as Indonesian and Sundanese~\cite{putri2024can}. While these models produce linguistically accurate text, they struggle to represent deeper cultural elements. This shortcoming is compounded by the predominance of Western-centric values in training datasets, as highlighted by Kharchenko \textit{et al.}~\cite{kharchenko2024well}, leading to cultural insensitivity and oversimplified stereotypes.

To assess cultural competency, researchers have developed benchmarks like BLEND, which includes 52.6k question-answer pairs spanning 16 countries and 13 languages, including low-resource languages such as Amharic and Sundanese~\cite{myung2024blend}. The results show that LLMs perform well for cultures that are well-represented online but struggle significantly with low-resource languages and nuanced cultural knowledge from underrepresented regions.

Additional frameworks, such as Hofstede’s Cultural Dimensions~\cite{bhatt2024extrinsic} and Bloom’s Taxonomy~\cite{chang2024benchmarking}, have also been used to evaluate LLMs' recall of factual cultural knowledge. However, these assessments reveal that LLMs consistently fall short in tasks requiring deeper cultural sensitivity, such as providing culturally appropriate advice or creatively adapting content. In contrast, human evaluators excel at these tasks, demonstrating an intuitive ability to adjust to cultural contexts~\cite{dudy2024analyzing, chiu2024culturalteaming}.

Although some methods have been proposed to improve cultural adaptation by dynamically incorporating external information during content generation~\cite{chang2024benchmarking}, significant gaps remain. These gaps are especially evident in tasks requiring profound cultural understanding, particularly for underrepresented groups. Bridging these gaps will require advancements in training data diversity, more inclusive benchmarking, and techniques for adaptive content generation to better align with cultural nuances.

\subsection{Summary}
Overall, LLMs have made great advances in cognitive, perceptual, analytical, emotional, social, and cultural competencies. They are skilled at structured reasoning, pattern recognition, data analysis, and creativity. Yet, they fall short in handling multi-step logic, real-time learning, genuine empathy, and complex social dynamics. Their ability to adapt to cultural nuances remains limited, often defaulting to stereotypes for underrepresented groups. While effective in controlled environments and structured tasks, LLMs struggle in unpredictable, real-world scenarios and continue to rely on external validation for consistency. Our Figure~\ref{fig:llm_competency_bar} provides a more detailed summary of the competency comparison between LLMs and human. Bridging these gaps will require significant enhancements in real-time learning, contextual understanding, and emotional awareness to more closely align LLMs with human-like adaptability and intelligence.

% Overall, LLMs have made considerable progress across cognitive, perceptual, analytical, emotional, social, and cultural skills, yet they still struggle to match human flexibility and depth. In cognitive tasks, they excel in structured reasoning and memory retention but face challenges with multi-step logic and real-time learning. Perceptually, LLMs have improved in pattern recognition and spatial awareness in controlled settings but often fail in dynamic, real-world situations. Their analytical skills show strengths in data interpretation and creativity, though they lack the intuition and emotional understanding that drives true innovation. LLMs have also advanced in emotional intelligence, recognizing and responding to emotions better than before, yet they still lack genuine empathy and deeper insight. Socially, LLMs demonstrate better interpersonal communication and collaboration skills but have difficulty fully understanding complex social dynamics and adapting to unpredictable contexts. In cultural knowledge, while they perform well with widely recognized cultures, they struggle with sensitivity toward underrepresented ones, often defaulting to stereotypes. Their planning and adaptability have improved, enabling them to manage structured tasks and create logical plans, though they still rely on external checks for consistency. Bridging these gaps will require enhancing real-time learning, contextual understanding, and emotional awareness to make LLMs more similar to human intelligence.

\section{LLMs in Human-Centric Studies}\label{sec:studies_improve_llm}

\begin{figure}
    \centering
    \includegraphics[width=1.0\linewidth]{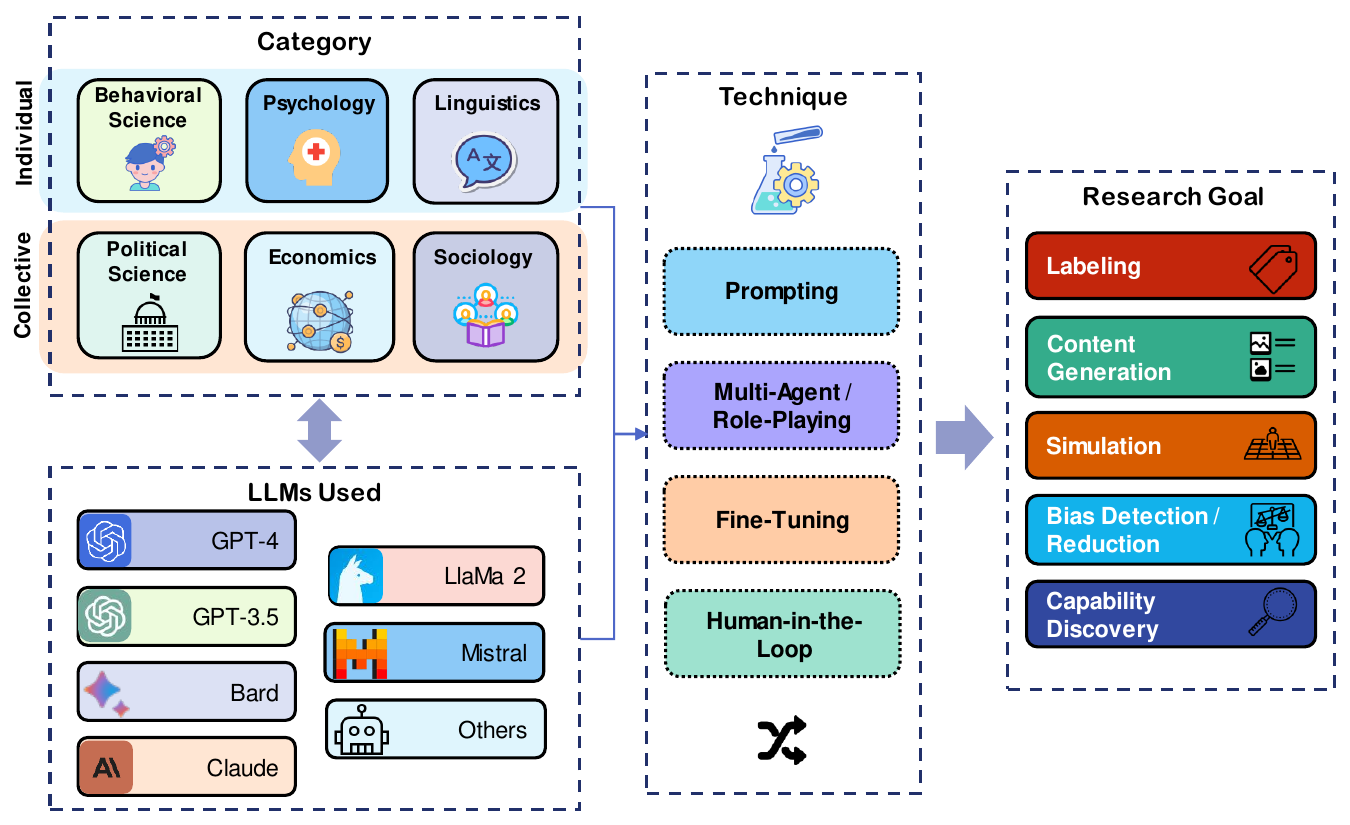}
    \caption{The conceptual framework for section \ref{sec:studies_improve_llm}, showing the human-centric domains that have been focused on using LLMs as a primary tool for investigation, the common LLM models that are employed, the most frequently seen techniques for using those models, which go from simple prompting to multi-agent systems and fine-tuning, or a combination of such methods, and finally, the research goals.}
    \label{fig:humanity_improving_conceptual}
\end{figure}

In the following section, we transition from evaluating LLMs in isolated tasks to exploring their application in real-world studies where humans are central to the research. Specifically, we focus on how LLMs perform in domains at two scales: individual, where an LLM performs tasks typically done by a single human, and collective, where multiple LLMs collaborate to achieve dynamic, group-based outcomes. Individual domains include behavioral science, psychology, and linguistics, whereas collective domains are composed of economics, and political science, and sociology. These studies aim to assess LLMs' ability to replicate human-like reasoning and interaction in a variety of real-world scenarios. Furthermore, the methodologies used in these studies can be categorized into three main approaches and are laid out in Figure \ref{fig:humanity_improving_conceptual}: (1) basic prompting, which involves sequentially querying a model to generate responses that build a broader understanding; (2) multi-agent prompting, where several LLMs interact autonomously based on predefined rules and theories; (3) fine-tuning, which involves retraining models with additional data to improve their performance in specific domains; and (4) human-in-the-loop, which involves a real person who interacts with the agents by inputting a stimulus into the system. These methods are often combined to create more human-like behaviors in LLMs, with frameworks drawing from established theories such as game theory, theory of mind, social learning theory, etc. In examining these applications, we consider various research goals of LLM-based contributions, including simulation, text content labeling and generation frameworks, methods for quantification and reduction of bias, and works that aim to discover new LLM human-centric capabilities. The following sections delve into the major works in this area, organized by their respective domains and sub-fields. The cited works and their respective contributions are also summarized in Appendix Tables \ref{tbl:econ_socio_psych_ling} and \ref{tbl:behavioralpolitical}.

\subsection{Individual Domains}

\begin{figure}
    \centering
    \includegraphics[width=0.85\linewidth]{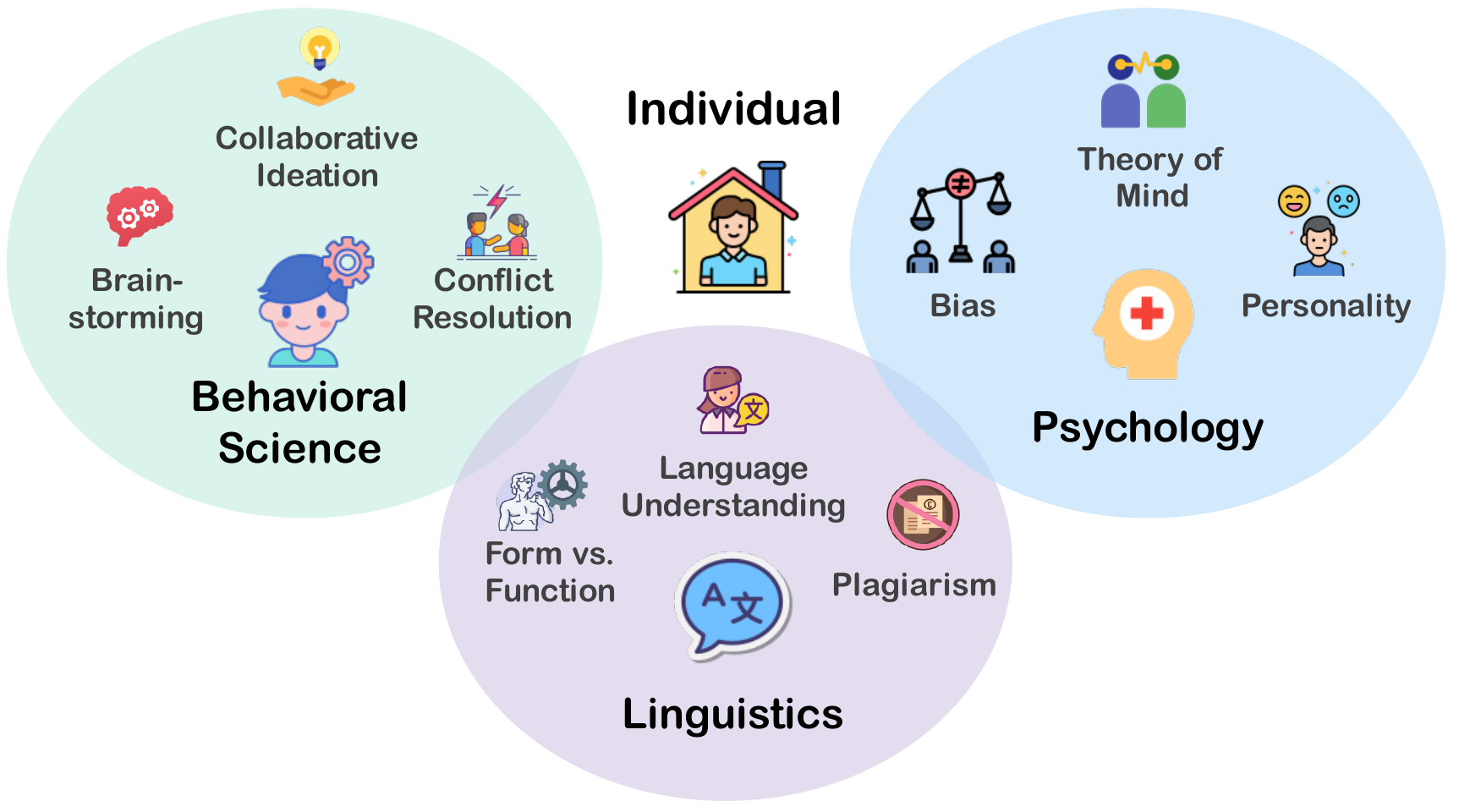}
    \caption{The individual domains consist of behavioral science, psychology, and linguistics, and each domain contains studies across a wide array of research problems from collaborative ideation and theory of mind, to plagiarism and conflict resolution, among several others.}
    \label{fig:indiv_domains}
\end{figure}

In this section, we provide an overview of the existing works that are human-centric with an individual emphasis, focusing on scenarios and applications where the LLM would be put to use in the capacity of an individual human or in small-scale group scenarios where emulating individual human behaviors can lead to realistic outcomes. The domains covered in this section include behavioral science, psychology, and linguistics, and the specific topics are shown in Figure \ref{fig:indiv_domains}.

\subsubsection{Behavioral Science}
Behavioral science is the study of human actions, decision-making processes, and interactions, drawing from fields like psychology, sociology, and cognitive science and aims to understand how individuals and groups behave in various settings, including social, organizational, and economic environments \cite{glanz2015health}. The knowledge from behavioral science provides valuable insights into the mechanisms behind cooperation, collaboration, and social learning, making it ideal for use in LLMs. 

Several studies explore how LLMs can be enhanced by incorporating them into collaborative environments \cite{he2024ai, shaer2024aiaugmented, guo2024embodied}. He \textit{et al.} \cite{he2024ai} investigate how LLMs can effectively participate in collaborative work scenarios, aiming to improve LLMs' ability to engage in collaborative ideation, whereas Shaer \textit{et al.} \cite{shaer2024aiaugmented} explore how LLMs can assist in collaborative writing, aiming to refine LLMs’ ability to generate creative and coherent content in teamwork settings. The previous two works use a similar virtual whiteboard design that allows agents to brainstorm together and come up with optimal ideas through mutual sharing. Similarly, Liu \textit{et al.} \cite{liu2024peergpt} delve into how LLMs can augment collaborative learning environments for children by designing scenarios where LLMs were included as team moderators and participants and determining effectiveness in fostering creative thinking among children. Wang \textit{et al.} \cite{wang2024unleashing} adds a novel element by examining cognitive synergy in multi-agent LLM systems through a novel self-collaboration agent design where personas with different strengths can work together to solve problems more effectively. The above works present important methods for understanding and improving LLMs' problem-solving capabilities through collaboration.
On a larger scale, Guo \textit{et al.} \cite{guo2024embodied} design a framework consisting of embodied agents and prompt-based organization structures to investigate the dynamics between teams and management. They find that with certain combinations of team composition, LLMs can optimize decision-making processes within organizational structures.

Some works also employ a simulation framework to discover how LLMs can interact in groups like humans do. Park \textit{et al.} \cite{park2023generative} create a community of LLM agents, each with unique characteristics and needs, to study social dynamics in an open-world game-like scenario. The agents are instantiated with a complex framework for realistic behavior, including an experience synthesis and memory architecture. Their work takes a bold step in exploring how LLM can be used to simulate complex human-like behaviors. CAMEL \textit{et al.} \cite{li2023camel} use a method they dub inception prompting in combination with role-playing to explore LLMs' ability to cooperate autonomously, leveraging theory of mind to enhance their collaborative potential. Inception prompting is a recursion-like method where the LLM assistant and agent prompt each other in a constant loop, where the prompts include tasks, roles, communication protocols, etc. Role-playing is a crucial part of the prompting technique, as it allows for the creation of a large number of varied conversations, namely 50,000. 

Shaikh \textit{et al.} \cite{shaikh2023rehearsal} and Park \textit{et al.} \cite{park2022social} develop multi-agent simulation scenarios to explore LLM's ability to resolve conflicts and simulate social media dynamics, respectively. Specifically, Shaik \textit{et al.} design a system for testing conflict resolution using role-based agents and an interlocutor to help them resolve a series of conflicts, guided by social science theory. They find that their simulation tool serves as an effective tool for real conflict resolution in the workplace. On the other hand, Park \textit{et al.} design a social media simulator using GPT-3. They infuse the system with a set of community guidelines and agents with specific roles. The results shed light on the way social media dynamics can be regulated with effective community design, providing insights to platform designers. 

By incorporating behavioral theories, these works provide valuable insights into how LLMs can better understand and simulate human interactions at different scales and in different scenarios, leading to improved cooperation mechanisms and collaborative problem-solving. The works also pave the way for future research on more sophisticated frameworks for modeling human-like behaviors in AI systems and exploring the long-term effects of LLM-augmented collaboration.

\subsubsection{Psychology}
Psychology investigates the mental processes and behaviors of individuals as they relate to their biological, cognitive, emotional, and social factors \cite{gerrig2015psychology}.
These insights into human thought and behavior are critical for refining LLMs to exhibit more human-like reasoning and empathetic engagement.
By incorporating psychological principles, LLMs can develop a deeper understanding of user emotions and intentions, leading to improved interactions that feel more personalized, emotionally intelligent, and aligned with human cognitive patterns.

Within this field, substantial of works investigated the approaches for improving the capability of LLMs in assisting psychological research.
For example, both Dillion \textit{et al.}~\cite{dillion2023can} and Demszky \textit{et al.}~\cite{Demszky2023} discuss various protocols for applying LLMs into psychological research, working as substitutions for the human volunteers who are studied, as well as the reliability of such approaches.
On the other hand, various studies explore the approaches for improving LLMs from the perspective of human psychology phenomena~\cite{strachan2024testing,stella2023using,chuang2024wisdom,zhang2023exploring}.
Among these studies, Strachan \textit{et al.}~\cite{strachan2024testing} explore whether LLMs can produce behavior comparable to human behavior through the lens of theory of mind, highlighting 
the importance of systematic approaches to enhance the capability of LLMs in  to align with human mentalistic inference processes. Both Chuang \textit{et al.}~\cite{chuang2024wisdom} and \textit{et al.}~\cite{zhang2023exploring} studied the collective intelligence, which is an important mechanism in human collaboration, among multiple LLM agents.
They shed light on an important way to improve LLM capabilities, namely by organizing multiple LLM agents to collaborate like humans.

In summary, research in this domain explores how LLMs can be integrated into psychological studies and improved through insights from human psychological phenomena.
These works all contribute to better understandings of the working mechanism of LLMs from a psychological perspective, shedding light on various possible approaches that may improve the capability of LLMs to emulate humans.

\subsubsection{Linguistics}
Linguistics investigates the structure, evolution, and use of language, analyzing how meaning is constructed and communicated across different languages and contexts \cite{yule2022study}.
% It also explores the cognitive mechanisms underlying language acquisition and processing.
By embedding linguistic theories in LLMs, models can better comprehend details in syntax and semantics of language, which allows for more accurate and contextually relevant language generation.
This linguistic foundation is key to improving the LLM's ability to handle multilingual tasks and adapt to varying communicative styles across cultures.

Since natural language is the major interface for LLMs to interact with humans, substantial works in this domain focus on uncovering the linguistic shortcomings in LLMs, and thereby providing guidance for further improving their capabilities~\cite{mahowald2024dissociating,lai2023chatgpt,pavlick2023symbols}.
For example, Mahowald \textit{et al.}~\cite{mahowald2024dissociating} evaluate LLMs by distinguishing between formal linguistic competence, which represents knowledge of linguistic rules and patterns, and functional linguistic competence, which requires understanding and using language in practical contexts.
Their findings show that while LLMs perform well in formal competence, their abilities in functional tasks are inconsistent. 
Based on such discovery, they point out that possible solutions for improving this aspect of capability include specialized fine-tuning and integrating external modules.
Also, Lai \textit{et al.}~\cite{lai2023chatgpt} evaluate ChatGPT’s performance beyond English.
Testing with multilingual NLP tasks, which cover 37 diverse languages including English, French, Spanish, Chinese, and etc., their findings reveal that ChatGPT performs significantly better in English compared to many other languages, particularly for tasks requiring complex reasoning.
Their study calls for further research that develop models with better multilingual understanding capability, where diverse data and language-specific fine-tuning are necessary.
Moreover, Pavlick \textit{et al.}~\cite{pavlick2023symbols} explore the potential of LLMs as models for human language understanding, highlighting two key factors that challenge their plausibility: the absence of symbolic structure and the lack of grounding.

The above works highlight the limitations of LLMs in linguistic competency, thereby emphasizing % bad functional linguistic competencies, weak multilingual performance, and lack of grounding.
the necessity for addressing these gaps with diverse training data, fine-tuning designs, and external modules.

% All of these shed light on the future direction of developing more plausible and reliable LLMs.

\subsection{Collective Domains}

\begin{figure}
    \centering
\includegraphics[width=0.92 \linewidth]{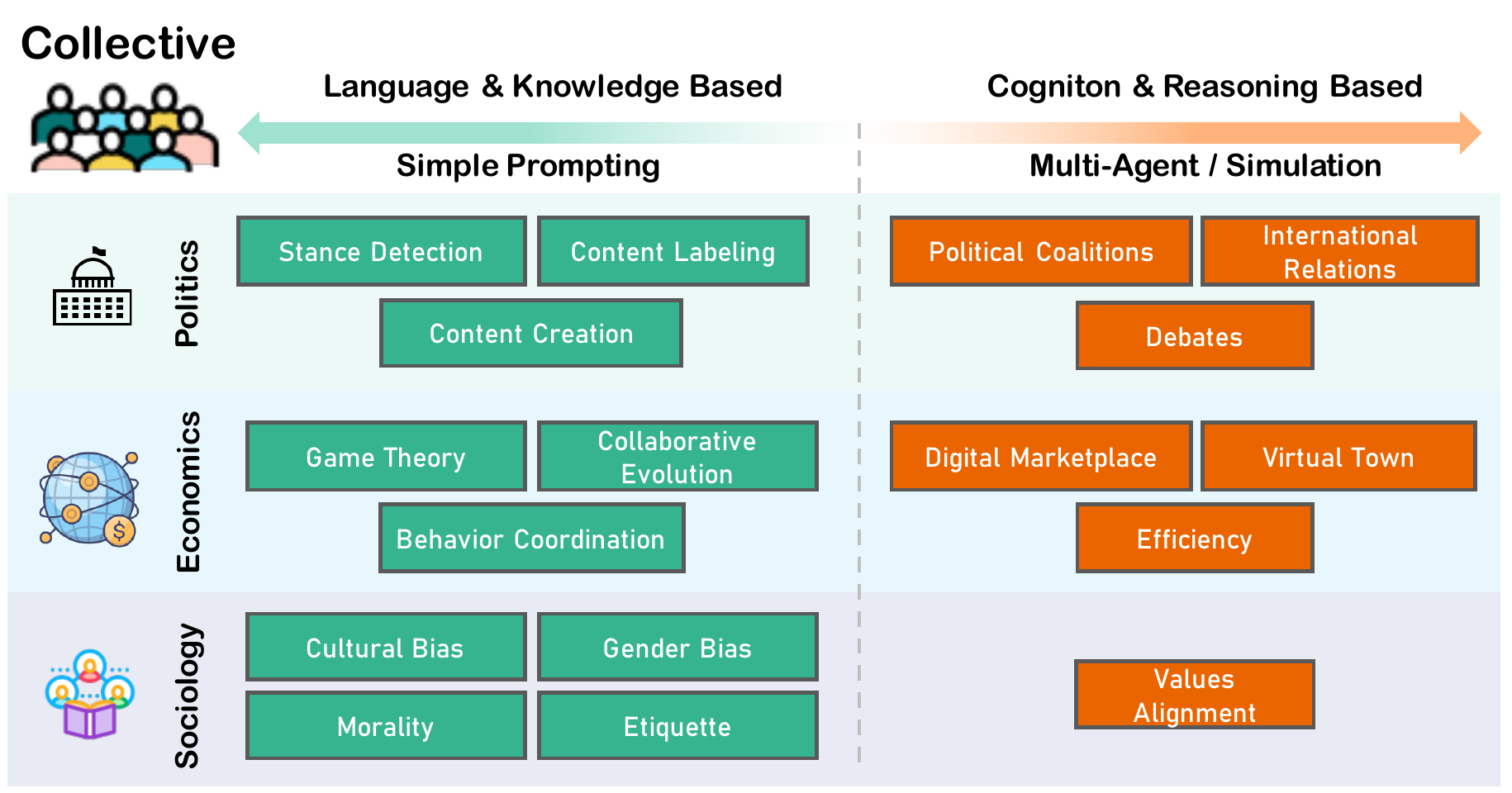}
    \caption{The collective domains consists of politics, economics, and sociology, and vary from more simple employment methods such as prompting, relying on language comprehension and knowledge retrieval, all the way to more complex multi-agent and simulation systems which rely on LLMs advanced abilities in cognition and reasoning.}
    \label{fig:domains_collective}
\end{figure}

In this section, we provide an overview of the existing works that are human-centric with a collective emphasis, focusing on scenarios and applications where LLMs are infused with theory and applied to test their ability to emulate realistic macro trends and behaviors. The domains covered in this section include politics, economics, and sociology, and the specific topics are shown in Figure \ref{fig:domains_collective}.

\subsubsection{Political Science}
Political science examines the ways in which governments allocate resources and make decisions, as well as the perceptions and behaviors of constituents in relation to bodies of authority \cite{shively2022power}. 
% manage  structures, behaviors, and dynamics of political systems and provides insights into how individuals and institutions shape and are shaped by ideologies and power structures. 
By leveraging knowledge from political science, LLMs can improve their ability to understand political discourse, predict political behaviors, and simulate complex political scenarios. Concepts like historical determinism, coalition formation, and the echo chamber effect offer pathways for LLMs to enhance their ability to model political phenomena, reduce bias, and generate more nuanced political analyses.
% By incorporating these theories, LLMs can be fine-tuned to better simulate political debates, tailor persuasive political messages, and detect biases, leading to more reliable and accurate political applications.

There are many recent studies that utilize LLMs to simulate political phenomena, with the aim of improving LLMs' ability to understand and predict political behaviors. For example, Hua \textit{et al.} \cite{hua2024war} examine how multi-agent LLM simulations can replicate large-scale historical events such as world wars. Using an architecture consisting of country agents, secretary agents, board, and stick, they reveal some of the factors that lead up to the outbreak of a conflict. Reflecting the components of a board game like risk, the board represents the inter-relationships between countries whereas the stick represents the internal record keeping mechanism for each country. This work represents a first step to understanding LLM's ability to model and replicate complex, collective human behaviors on a societal scale. On a slightly smaller scale, Moghimifar \textit{et al.} \cite{moghimifar2024modelling} explore how LLMs can uncover the facts that lead to the formation of political coalitions, which can prevent domestic conflicts. By presenting a dataset and a hierarchical Markov decision process tailored to this task, they open up an avenue for future LLM works to predict outcomes of political negotiations. Similarly, Taubenfeld \textit{et al.} \cite{taubenfeld2024systematic} simulate political debates using LLMs to understand biases in political discussions. They find that the evolution of views of the agents in a debate differs from real human behaviors, with both participants of the debate converge towards the models' inherent bias. They also find that fine-tuning can make a certain bias more pronounced. These findings shed light into the differences between LLMs and real human tendencies and can inform methods for reducing bias in LLM outputs.

Further directions of LLMs' application towards the political arena include political content generation and content labeling. Hackenburg \textit{et al.} \cite{hackenburg2024evaluating} and Goldstein \textit{et al.} \cite{goldstein2024how} study how LLMs can be improved to tailor political messages and generate convincing content, honing LLMs' ability to adapt to audience profiles. More specifically, Hackenburg \textit{et al.} develop a web-application that can integrate user-reported demographic information into prompts to generate indiviudally tailored political messaging, whereas Goldstein \textit{et al.} leverages a user survey to determine the persuasiveness of LLM generated so-called propaganda theses. 

Moreover, several studies address stance detection and misinformation evaluation using LLMs \cite{zhu2023cana, zhang2024llmdriven, lan2024stance, urman2023silence}. Zhu \textit{et al.} \cite{zhu2023cana} assess ChatGPT’s ability to replicate human-annotated labels across various social computing tasks, including political stance detection. Zhang \textit{et al.} \cite{zhang2024llmdriven} explore how LLMs can determine the relationship between explicit political standpoints and correlated views. Additionally, Li \textit{et al.} \cite{li2024mitigating} address bias in stance detection by introducing a calibration framework to help LLMs generate more balanced and accurate political analyses.

% Moreover, several works tackle the tasks of stance detection with LLM \cite{zhu2023cana, zhang2024llmdriven, lan2024stance} as well as evaluation LLMs for the presence of misinformation LLMs \cite{urman2023silence}. Zhu \textit{et al.} \cite{zhu2023cana} evaluate ChatGPT for its ability to reproduce human annotated labels on various social computing tasks including political stance detection, whereas Zhang \textit{et al.} \cite{zhang2024llmdriven} use LLM to attempt to determine the relationship between explicit political standpoints and their correlated views. Furthermore, Li \textit{et al.} \cite{li2024mitigating} go a step further in addressing the issue of bias in stance detection by introducing a calibration framework, which can help LLMs produce more balanced and accurate political analyses.

In sum, the described works demonstrate the application of LLMs in simulating political phenomena, while also exploring their potential in tailoring political messages and detecting political stances. These studies reveal both the capabilities and limitations of LLMs in modeling complex political behaviors, highlighting areas for improvement in reducing bias and enhancing the accuracy of political simulations and analyses.

\subsubsection{Economics}
Economics involves understanding how individuals, businesses, and governments make choices about the allocation of resources, production, and consumption of goods and services \cite{mankiw2020economics}. LLMs, which are increasingly being used to simulate economic behaviors and predict market outcomes, can benefit from the theoretical frameworks in the field of economics. 
% By integrating economic theories, LLMs can be enhanced to better simulate complex market dynamics, optimize decision-making in economic simulations, and reduce biases in economic predictions, thereby making them more useful in both academic research and practical applications.

LLMs are being explored for their ability to simulate economic behaviors and improve their understanding of complex market dynamics. In several existing works, LLMs are evaluated for their ability to emulate human behavior across a handful of economic scenarios, including game theory, macroeconomic trends, and market dynamics. Various works approach different aspects of game theory. A work by Guo \cite{guo2023gpt} represents an early attempt at using prompting and architectures to investigate how LLMs can be used in strategic game experiments like the ultimatum game and prisoner’s dilemma. In a related vein, Horton \textit{et al.} \cite{horton2023large} endows LLMs with various features such as information and preferences, determining that the outcome of LLMs in a set of classic economic scenarios reinforces prior works' conclusions, showing that LLMs can reliably simulate economic decision-making. Similarly, Akata \textit{et al.} \cite{akata2023playing} explore coordination and cooperation mechanisms in LLMs in a variety of prisoner's dilemma scenarios, finding consistently high performance and behavioral signatures among coordinating agents. Going even further, Suzuki \textit{et al.} \cite{suzuki2024evolutionary} simulate the evolution of human decision-making in economic scenarios, treating personality traits as "genes" in prisoner’s dilemma studies. Such studies can be used to enhance the way LLMs' model micro-level economic behaviors. In turn, Li \textit{et al.} \cite{li2024econagent} design an LLM architecture including memory, reflection, and decision modules in combination with macro-economic theory and heterogeneous decision-making mechanisms to explore how LLMs can model macro trends.

Finally, market dynamics are addressed by Weiss \textit{et al.} \cite{weiss2023rethinking}, who study information markets by simulating a digital marketplace, drawing on social learning theory and buyer's paradox. This work exploits the unique qualities of human cognition such as forgetting and assessing information to create more realistic agents. Meanwhile, Zhao \textit{et al.} \cite{zhao2024competeai} examine market competition by simulating a virtual town where LLM agents compete for customers. These works can help improve how LLMs agents with asymmetric information can better predict market dynamics.

In summary, recent research has explored the application of LLMs in simulating various economic scenarios, ranging from game theory experiments and macroeconomic trends to complex market dynamics. These studies demonstrate the potential of LLMs to model economic decision-making processes and predict outcomes in diverse contexts, while also highlighting areas for improvement in creating more realistic and nuanced economic simulations.

\subsubsection{Sociology}
Sociology examines the structures and dynamics of social life, social change, and the interactions between families as well as macro-scale human communities \cite{macionis2005sociology}. 
% , offering insights into how individuals and groups interact within various social frameworks
These studies provide a deeper understanding of social bias, ethical norms, and human morality.
By integrating sociological principles, LLMs can be prompted or fine-tuned to recognize and respect cultural contexts, making their responses more socially aware and ethically sensitive when addressing diverse populations.
This improves their ability to adapt to different societal expectations and enhances the realism of their interactions.

One major category of works in this field aims to enhance the social capabilities of LLMs from various aspects by comprehensively uncovering their shortcomings~\cite{scherrer2024evaluating,rao2024normad,dwivedi2023eticor,zheng2024ali}.
For example, Scherrer \textit{et al.}~\cite{scherrer2024evaluating} identify certain limitations in the moral decision-making processes of LLMs, particularly in ambiguous scenarios where the moral choice is unclear, pointing out the necessity of approaches, such targeted fine-tuning, to address these gaps.
Rao \textit{et al.}~\cite{rao2024normad} and Dwivedi \textit{et al.}~\cite{dwivedi2023eticor} highlight deficiencies in LLMs' adaptability to different cultural contexts and regional etiquettes.
These stress the need of incorporating culturally diverse data and region-specific training protocols to improve these capabilities.
Also, Zheng \textit{et al.}~\cite{zheng2024ali} develop a multi-agent framework to test the misalignment of LLMs with human values through scenario generation and refinement, suggesting a basis for continuous enhancement in value alignment.
On the other hand, other researchers investigated using human-in-the-loop approach to improve the social capability of LLMs~\cite{tao2023auditing,duan2023denevil}.
For instance, Tao \textit{et al.}~\cite{tao2023auditing} explore cultural bias for five commonly used LLMs, and design cultural prompting as a control strategy to increase cultural alignment of LLMs with specific country or territory.
Duan \textit{et al.}~\cite{duan2023denevil} design a framework that dynamically generates ethical prompts and evaluates model responses based on human-defined moral values, refining the LLMs with finetuning.
% Moreover, there existing researchers studied the social features of LLMs via simulation approaches.
% For example, Kotek \textit{et al.}~\cite{duan2023denevil} use LLM agents to simulate people in the society, looking into the stereotypical gender bias in LLMs.

In summary, the works in this field primarily focus on identifying and addressing the social limitations of LLMs.
Having uncovered the limitations, researchers propose solutions like fine-tuning, incorporating diverse data, and using human-in-the-loop approaches to enhance the social and ethical capabilities of LLMs.

\subsection{Summary} 

Overall, the research in human-aware fields for improving LLM has validated advanced and often very realistic behaviors, decision-making, and thought processes being made by LLMs using a variety of unique and complex architecture paradigms. From the findings of such works, it is evident that the emergence of LLMs poses great implications for all aspects of society, from politics, to economics, to psychology and linguistics, among others, in turn shedding light on the limits of the capabilities of pre-trained generative models. The findings, thereby, can be used inform both the academic community and decision makers, both at a corporate and governmental level, how to adapt relevant systems of society to adapt to this groundbreaking new development in the way information is sought and generated. Furthermore, these findings can serve as a feedback stream to LLM developers to better mold their LLM outputs to users' and society's needs. However, the road forward is not entirely clear, as users of LLMs and society at large may have contradictory aims. As such, there is a need for an even deepening and broadening body of research on the capabilities of LLMs so as to uncover their functions that can be most conducive for positive human-centered outcomes. As the research becomes more nuanced and comprehensive, the answers may become more clear. 

\section{Open Challenges and Future Directions}
\label{sec:chall}

% As LLMs continue to show promise in various fields, particularly within the humanities and social sciences, several critical challenges remain unresolved. In this section, we separate these challenges into two key areas—improving core LLM competencies (Section 3) and enhancing their integration in human-centric studies (Section 4)—and discuss potential future directions for addressing them.

While LLMs continue to demonstrate impressive advancements, several critical challenges remain that require further attention. In this section, we present the challenges within two main areas: (1) Human-Centric Evaluation of LLMs and (2) LLMs in Human-Centric Studies. The former focuses on identifying key skill areas where LLMs need improvement, such as their adaptability to real-world contexts, ability to understand and respond to human emotions, and sensitivity to cultural dynamics. The latter explores how LLMs can be effectively applied within individual and collective domains. We also outline potential directions for addressing these challenges.

% The interaction of LLMs in the humanities is a highly promising field that has gained considerable attention in recent years, resulting in significant achievements across disciplines such as sociology, psychology, and political science. Despite these successes, several unresolved issues and challenges remain that need to be addressed. In this section, we will explore these open problems and discuss potential future directions.

% \subsection{Human-Centric Evaluation of LLMs}
\subsection{Advancing Real-Time Learning and Adaptability in LLMs}
\textbf{Open Discussion:} One significant challenge LLMs face is their inability to perform real-time learning and adapt to new information dynamically. While they excel in structured environments and can generate coherent plans, LLMs do not have the capability to update their parameters based on new data or experiences during inference. This limitation hinders their ability to evolve and improve in response to changing circumstances or user interactions, which is crucial for applications that require continuous learning and adaptability.\\

\noindent \textbf{Future Direction:} A potential direction to address this challenge is the development of hybrid learning architectures that combine LLMs' static pre-trained knowledge with online learning algorithms. Integrating reinforcement learning techniques, such as RLGF, could allow LLMs to adapt dynamically in real-world scenarios by continuously refining their decision-making processes based on real-time feedback. This approach would enable LLMs to make context-aware adjustments in unpredictable environments and better simulate human cognitive flexibility.

\subsection{Enhancing Emotional Intelligence and Empathy in LLMs}

\textbf{Open Discussion:} While LLMs have advanced in recognizing emotions, their capacity to express genuine emotional intelligence and empathy remains limited, particularly in nuanced human interactions. Despite progress in identifying and interpreting emotions, they often lack the depth required to generate responses that truly resonate with individuals' emotional states across diverse contexts. As a result, their interactions can come across as mechanical or overly neutral, missing the subtlety needed for empathetic communication. This limitation affects their effectiveness in sensitive areas like mental health support, customer service, and human-computer interaction, where a more human-like empathetic response is crucial.\\

\noindent \textbf{Future Direction:} Looking forward, enhancing the emotional intelligence of LLMs requires a multi-faceted approach that goes beyond basic emotional recognition. A key strategy involves integrating affective computing techniques to enable models to interpret and respond to emotional cues in text and speech with greater accuracy. Additionally, employing reinforcement learning strategies focused on emotional reinforcement can help fine-tune the model's responses based on real-time user feedback in emotionally charged interactions. Incorporating culturally diverse real-world data into LLM training is also essential to improve empathetic accuracy, ensuring these models become more sensitive to the emotional expressions unique to different cultures. By advancing these capabilities, LLMs can play a more meaningful role in applications that require human-like emotional support and culturally aware communication.

\subsection{Improving Cultural Competency and Reducing Bias in AI Interactions}
\textbf{Open Discussion:} While LLMs have shown advancements in their cultural competency, they still struggle to understand and respond accurately to nuanced cultural contexts, particularly for underrepresented or minority cultures. This limitation often leads to generalized or stereotypical outputs that fail to capture the subtleties of different cultural perspectives. The challenge lies in making LLMs more culturally adaptive to provide responses that respect and reflect diverse societal norms and values.\\

\noindent \textbf{Future Direction:} To address the issues, a possible direction is to develop more inclusive training datasets that better represent a wide range of cultural and linguistic contexts. Additionally, techniques like Retrieval-Augmented Generation (RAG) can be leveraged to dynamically incorporate relevant cultural knowledge from external databases during interactions, enhancing the accuracy and sensitivity of LLM responses. Future research should focus on creating adaptive models that can better handle context-specific nuances, promoting culturally aware interactions and reducing unintended biases in diverse global settings. This approach aims to improve the reliability and inclusivity of LLMs when engaging with users from various cultural backgrounds.

\subsection{LLMs in Human-Centric Individual Domains}

\textbf{Open Discussion:}
In the context of individual domains such as behavioral science, psychology, and linguistics, LLMs face several key challenges. In behavioral science, LLMs struggle with accurately simulating human social learning and collaboration in dynamic, real-world scenarios. In psychology, LLMs have difficulty consistently emulating human cognitive and emotional processes, often failing to adapt to evolving user emotions and intentions. Additionally, in linguistics, LLMs have limited capabilities in handling deep semantic understanding and pragmatic language use across diverse contexts, especially in multilingual settings. These limitations hinder LLMs from fully replicating the depth of human cognition, emotional nuance, and cultural sensitivity required for authentic, human-like interactions.\\

\noindent \textbf{Future Direction:}
To address these challenges, future research should focus on developing hybrid LLM architectures that integrate real-time feedback, psychological models, and external linguistic knowledge bases. Specifically, reinforcement learning techniques and multi-agent systems could enhance LLMs' ability to simulate human collaboration and social behaviors more effectively. In psychology, incorporating more detailed cognitive and emotional models into LLM training could improve the models' ability to generate adaptive, empathetic responses. For linguistics, improving multilingual performance through diverse training datasets, fine-tuning for specific languages, and coupling LLMs with symbolic reasoning systems can lead to more contextually accurate and culturally aware language generation. These improvements would enable LLMs to better navigate complex interactions in human-centric applications.

\subsection{LLMs in Human-Centric Collective Domains}
\textbf{Open Discussion:}
In collective domains such as political science, economics, and sociology, LLMs face significant challenges in accurately simulating complex systems and predicting long-term behaviors. In political science, LLMs struggle with modeling intricate power dynamics, political alliances, and the outcomes of political negotiations. In economics, the models often fail to simulate complex market behaviors and economic decision-making, especially when dealing with irrational actors or unpredictable external factors. Additionally, in sociology, LLMs have difficulty modeling the complexities of social structures, ethical norms, and cultural dynamics, often reinforcing existing biases and stereotypes. These challenges limit the ability of LLMs to provide reliable and unbiased insights in large-scale human-centric studies.\\

\noindent \textbf{Future Direction:} 
To overcome these limitations, future research should focus on enhancing the integration of domain-specific frameworks into LLM architectures. In political science, incorporating game theory, coalition analysis, and real-time political data streams could improve the models' ability to simulate political behaviors and predict outcomes. In economics, coupling LLMs with macroeconomic models and integrating behavioral economics principles would enable more realistic simulations of both micro- and macro-level economic trends. For sociology, building more inclusive training datasets and employing human-in-the-loop approaches can help LLMs generate more socially and ethically aware responses. These advancements will improve the accuracy and cultural sensitivity of LLMs when applied to complex human-centric studies in collective domains.

\section{Conclusion}
\label{sec:Con}

In this paper, we take the pioneering step to treat LLMs as entities with human-like intelligence and systematically evaluate them through a humen-centric framework. 
We first present a comprehensive summary of the intersection between AI and the humanities, contextualizing the development of LLMs within broader interdisciplinary studies. Next, we elaborate on the evaluation of LLMs in key cognitive, perceptual, analytical, and social domains, bridging the gap between theoretical research and practical applications. We then emphasize interdisciplinary collaboration to improve LLMs future capabilities of understanding human mind and behavior, followed by a discussion of remaining open challenges and future research. We hope that our summaries and compilation of resources can be useful for researchers who wish to further uncover the abilities of LLMs to come into contact with our lives in various complex ways, and to improve the design and function of LLMs to act in a manner more consistent with human needs.

\bibliographystyle{unsrt}
\bibliography{Ref} \clearpage
\section{Appendix}
\begin{table}[H]
\centering
\newcolumntype{L}[1]{>{\raggedright\arraybackslash}p{#1}}
\newcolumntype{Y}{>{\raggedright\arraybackslash}X} % New column type for text wrapping
\scriptsize 
\vspace{-5px}

\caption{Evaluation of LLM Capabilities Across Cognitive, Perceptual, and Analytical Domains}
\vspace{-5px}

\resizebox{\textwidth}{!}{%
\begin{tabular}{p{1.2cm}p{0.6cm}p{0.6cm}p{3cm}p{6.5cm}p{3.3cm}p{1.3cm}}
% {@{}cccccccc@{}

\toprule
\textbf{Skill}  & \textbf{Paper} & \textbf{Year} & \textbf{LLMs} & \textbf{Highlight} & \textbf{Evaluation Metrics}& \textbf{Compares with Humans}
  \\ \midrule
\multirow{3}{*}{\makecell[{{l}}]{Reasoning}} 

&   \cite{liu2023logiqa}             & 2023         & GPT-2, GPT-3                           & Introduces LogiQA 2.0 dataset for evaluating LLM’s capabilities in complex logical reasoning &   Accuracy, precision, recall, and F1-score   &   \ding{51} \\ \cmidrule(l){2-7} 
&   \cite{saparov2022language} & 2022 & InstructGPT, GPT-3 & Creates PrOntoQA, a first-order logic-based synthetic dataset designed to analyze LLMs’ reasoning abilities by converting their chain-of-thought prompts into symbolic proofs & Strict to more relaxed version of proof accuracy & \ding{55} \\ \cmidrule(l){2-7}
&   \cite{talmor2022commonsenseqa} & 2022 &  T5, UNICORN, GPT-3 & Designs CommonsenseQA 2.0, a gamification-based benchmark, for exploring AI common sense limits & Accuracy, consistency & \ding{51} \\  \cmidrule(l){2-7} 
&   \cite{onoe2021creak} & 2021 &  T5-3B & Presents Creak, a dataset for commonsense reasoning on entity knowledge & Accuracy & \ding{51} \\  \cmidrule(l){2-7} 
 &    \cite{han2024inductive}             & 2024         & GPT-3.5, GPT-4                           &       Examines inductive reasoning in humans and LLMs using property induction tasks & Argument strength, sign tests, correlation, phenomenon capture
  &   \ding{51}                                                          \\ \cmidrule(l){2-7} 

&   \cite{wang2024causalbench} & 2024 &  LLaMA, OPT, InternLM, Falcon, GPT-3.5-Turbo, GPT-4, GPT-4-Turbo, etc. & Proposes CausalBench benchmark for measuring LLMs’ causal learning capability & F1 score, structural hamming distance, structural intervention distance & \ding{55} \\ \cmidrule(l){2-7}
&   \cite{hua2024disentangling} & 2024 & Qwen, LLaMA, GPT-3.5 & Presents ContextHub to assess LLMs' reasoning across abstract and contextual tasks & Average F1 score & \ding{51} \\ 
\cmidrule(l){2-7} 
&   \cite{cobbe2021training} & 2021 & GPT-3 & Designs GSM8K, a dataset of grade school math word problems, that evaluates LLMs’ mathematical reasoning capabilities & Test solve rate & \ding{55} \\  
\cmidrule(l){2-7} 
&   \cite{suzgun2022challenging} & 2022 & PaLM, Codex (code-davinci-002) & Showcases that CoT prompting significantly enhances LLMs’ performance on the challenging BIG-Bench Hard tasks & Exact match & \ding{51} \\  
 \midrule

 \multirow{3}{*}{\makecell[{{l}}]{Learning}} 

&   \cite{chang2023learning}& 2023& GPT-4, Claude2, LLaMA2 & Utilizes CommonGen tasks to measure the learning ability of LLMs & BLEU, CIDEr-D, SPICE& \ding{51} \\ \cmidrule(l){2-7}
&   \cite{an2023learning}& 2023& LLaMA, GPT-4, LLaMA, LLaMA2, WizardMath, MetaMath, CodeLLaMA & Introduces LEMA (Learning from Mistakes), a method to improve LLM reasoning by learning from mistakes, using error-correction data pairs for fine-tuning & Accuracy & \ding{51} \\ \cmidrule(l){2-7}  
&   \cite{jovanovic2024towards}& 2024& BERT, GPT-4 & Investigates incremental learning approaches including continual learning, meta-learning, parameter-efficient learning, etc. & Accuracy & \ding{51} \\ 
% \cmidrule(l){2-7}  
% &   \cite{ren2024learning}& 2024& Pythia, T5-small, T5-large, LLaMA-2, RoBERTa & LLMs' learning abilities are tested using datasets such as Antropic HH and UltraFeedback. & Alignment performance, Generalization ability, Squeezing Effect & \ding{55} 
\midrule
                          
\multirow{3}{*}{\makecell[{{l}}]{Pattern \\ Recognition}} 
& \cite{karamcheti2023language}& 2023 & GPT-3.5 & Introduces a framework for language-driven visual representation learning for robotics & Various metrics across five robotics tasks & \ding{55} \\ \cmidrule(l){2-7}
& \cite{qi2024shapellm}& 2024 & LLaMA & Presents a 3D multimodal LLM designed for embodied interaction, achieving state-of-the-art performance in 3D geometry understanding and language-unified interaction tasks & Fine-tuned accuracy, zero-shot accuracy on various benchmarks & \ding{55} \\ \cmidrule(l){2-7}
& \cite{hong20233d}& 2023 & GPT-4 & Proposes 3D-LLMs that incorporate the 3D physical world into LLMs, enabling models to perform 3D captioning, task decomposition, and spatially grounded dialogue & BLEU-1, CIDER, accuracy on 3D question answering benchmarks & \ding{51} \\ \cmidrule(l){1-7}

  \multirow{3}{*}{\makecell[{{l}}]{Spatial \\ Awareness}}
& \cite{tan2024towards}& 2024 & GPT-4o & Proposes CRADLE, a multimodal agent framework for general computer control, capable of operating diverse software and completing complex tasks in video games & Success rate across different software and games & \ding{51} \\ \cmidrule(l){2-7}
& \cite{wang2023robogen}& 2024 & GPT-4 & Automates the learning process of a generative robotic agent via generative simulation with minimal human supervision & Skill diversity, success rate in simulations & \ding{55} \\ \cmidrule(l){2-7}
& \cite{schumann2024velma} & 2024 & GPT-4, CLIP & Introduces an embodied LLM agent for vision and language navigation in Street View, achieving 25\% improvement in task completion over state-of-the-art baselines & Task completion rate in VLN & \ding{51} \\ \midrule
                          
\multirow{3}{*}{\makecell[{{l}}]{Data \\ Interpretation}}

& \cite{hong2024data} & 2024 & GPT-4 & Introduces a data science agent that integrates dynamic planning and tool utilization enhance performance on data-centric tasks & Machine learning accuracy, MATH dataset accuracy, open-ended task success & \ding{51} \\ \cmidrule(l){2-7}
& \cite{boiko2023autonomous}& 2023 & GPT-4 & Develops Coscientist, an AI system capable of autonomously designing, planning, and performing chemical experiments, advancing automated chemical research & Task success rate in experimental synthesis, optimization tasks & \ding{55} \\ \cmidrule(l){2-7}
& \cite{polak2024extracting}& 2024 & GPT-4 & Proposes ChatExtract, a conversational LLM-based method for automated data extraction from materials science literature via prompt engineering and redundancy & Precision, recall for data extraction & \ding{55} \\ \midrule

\multirow{3}{*}{\makecell[{{l}}]{Information \\ Processing}}
% & \cite{zhang2023large} & 2023 & GPT-3, BART & Introduces RPT (Relational Pre-trained Transformer), a Transformer-based denoising autoencoder that generalizes BERT and GPT architectures to automate data preparation tasks & F1 & \ding{55} \\ \cmidrule(l){2-7}
& \cite{borisov2022language} & 2022 & GPT-2 & Presents GReaT, a method that utilizes auto-regressive LLMs to generate highly realistic synthetic tabular data & Efficiency, distance to closest records histogram, ROCAUC, F1  & \ding{55} \\ \cmidrule(l){2-7}
& \cite{zou2024docbench} & 2024 & GPT-4, GPT-4o, GLM-4,  KimiChat, Mistral, LLaMA-2, LLaMA-3, etc.
% GPT-3.5, ,ChatGLM3, InternLM2 
& Designs DOCBENCH, a comprehensive benchmark designed to evaluate LLM-based document reading systems & Accuracy & \ding{51} \\ \cmidrule(l){2-7}
& \cite{dai2024cocktail} & 2024 & BERT, RoBERTa, DistilBERT, MiniLM, T5 & Proposes Cocktail, a benchmark of 16 datasets integrating human-written and LLM-generated content to evaluate information retrieval models in mixed-source data & Retrieval accuracy & \ding{51} \\ \midrule

 \multirow{3}{*}{\makecell[{{l}}]{Creativity}}     
& \cite{girotra2023ideas} & 2023 & GPT-4 & ChatGPT-4 outperforms students in creative generation, with high output and quality of ideas & Accuracy, Consistency, Relevance & \ding{51} \\ \cmidrule(l){2-7}
& \cite{xu2024jamplate} & 2024 & GPT-4 & Introduces Jamplate system design to enhance LLM's creative reflection ability & Accuracy & \ding{55} \\ \cmidrule(l){2-7}
% & \cite{li2024map} & 2024 & GPT, BERT & Presents a new mapping method for human LLM interaction mode & Human-AI Collaboration Dimension, Implement-Creation Dimension & \ding{55} \\ \cmidrule(l){2-7}
& \cite{lu2024llm} & 2024 & GPT-4, GPT-3.5-turbo & Proposes a framework and role-playing techniques to enhance LLM creativity, and use Alternative Uses Task to evaluate the creativity of LLM & Fluency, Flexibility & \ding{51} \\ 
% \midrule

\bottomrule
\label{table:paper_summary}
\end{tabular}
}
\end{table}

\begin{table}[ht]
\centering
\newcolumntype{L}[1]{>{\raggedright\arraybackslash}p{#1}}
\newcolumntype{Y}{>{\raggedright\arraybackslash}X} % New column type for text wrapping
\scriptsize 
\vspace*{-5px}

\caption{Evaluation of LLM Capabilities in Executive Functioning and Social Domains}
\vspace*{-5px}

\resizebox{\textwidth}{!}{%
\begin{tabular}{p{1.2cm}p{0.6cm}p{0.6cm}p{3cm}p{6.5cm}p{3.1cm}p{1.5cm}}
% {@{}cccccccc@{}

\toprule
\textbf{Skill}  & \textbf{Paper} & \textbf{Year} & \textbf{LLMs} & \textbf{Highlight} & \textbf{Evaluation Metrics}& \textbf{Compares with Humans}
  \\ \midrule

\multirow{3}{*}{\makecell[{{l}}]{Adaptability \\  and \\Flexibility}} 

&\cite{luu2024context}&2024&GPT-4& Proposes a framework combining LLMs, Stochastic Gradient Descent, and optimization-based control to help autonomous systems manage complex tasks with latent risks &Deviation squared, Sum of squared differences&\ding{55}\\\cmidrule(l){2-7}
&\cite{kim2023dynacon}&2023&GPT-3.5& Presents DynaCon, a system that uses LLMs for context-aware robot navigation in unknown environments without maps &Success Rate, Time-step, Number of Experiments&\ding{55}\\\cmidrule(l){2-7}
&\cite{shvartzshnaider2024llm}&2024&GPT-4, LLaMA-3.1 &  Introduces LLM-CI, an open-source framework to assess privacy norms encoded in LLMs using a multi-prompt methodology to address prompt sensitivity &Percentage of invalid responses&\ding{55}\\

\midrule

 \multirow{3}{*}{\makecell[{{l}}]{Planning\\ and \\ Organization}} 
& \cite{song2023llm} & 2023 & GPT-3 & Uses the ALFRED dataset to evaluate the planning and organizational capabilities of LLM & Success Rate, Goal-Condition Success Rate & \ding{55} \\ \cmidrule(l){2-7}
& \cite{kambhampati2024llms} & 2024 & GPT-4, GPT-3.5 & Leverages PlanBench benchmark to evaluate LLM's planning and organizational capabilities & Success Rate, Correctness Rate, Plan Pass Rates & \ding{55} \\ \cmidrule(l){2-7}
& \cite{gundawar2024robust} & 2024 & GPT-4 Turbo, GPT-3.5 Turbo & Implements Travel Planning Benchmark to evaluate LLM's planning and organizational capabilities & Delivery Rate, Final Pass Rate & \ding{51} \\ \cmidrule(l){2-7}
& \cite{sharan2023llm} & 2023 & GPT-4, LLaMA2, GPT-3 & Applies nuPlan Benchmark to evaluate LLM's planning and organizational capabilities & Score, Drivable & \ding{55} \\ 
\midrule
                          
\multirow{3}{*}{\makecell[{{l}}]{Decision \\ Making}}  
& \cite{huang2024far} & 2024 & GPT-3.5, GPT-4, Gemini-Pro, LLaMA-3.1, Mixtral, Qwen-2 & Introduces GAMA-Bench, a framework for evaluating LLM's decision-making abilities in multi-agent environments & Nash equilibrium & \ding{51} \\ \cmidrule(l){2-7}
& \cite{schramowski2022large} & 2022 & BERT, GPT-2, GPT-3 & Assesses the risks of relying on LLMs for reasoning under uncertainty & Toxicity probability, expected maximum toxicity & \ding{51} \\ \cmidrule(l){2-7}
& \cite{echterhoff2024cognitive} & 2024 & GPT-3.5-turbo, GPT-4, LLaMA-2  & Presents BIASBUSTER, a framework that evaluates and mitigates cognitive biases in LLM-assisted decision-making tasks & Euclidean distance & \ding{51} \\ \midrule

\multirow{3}{*}{\makecell[{{l}}]{Interpersonal \\Communication} }
& \cite{park2023generative} & 2023 & ChatGPT & Proposes generative agents as believable simulacra of human behavior, implementing them in a sandbox game environment to simulate social interactions, forming relationships, and executing plans autonomously & Observational consistency and emergent behavior analysis & \ding{55} \\ \cmidrule(l){2-7}
& \cite{shanahan2023role} & 2023 & GPT-4, LLaMA & Introduces a conceptual framework for understanding LLM-based dialogue agents as role-playing entities, emphasizing metaphorical approaches to avoid anthropomorphism in agent design & Qualitative analysis of role-play interactions & \ding{55} \\ \cmidrule(l){2-7}
& \cite{strachan2024testing} & 2024 & GPT-4, GPT-3.5, LLaMA2-70B & Examines the capability of LLMs to exhibit theory of mind by testing them on a comprehensive set of psychological tasks, showing mixed performance across tasks like false beliefs, irony, and faux pas & Performance accuracy on theory of mind tests & \ding{51} \\ \midrule
                          
\multirow{3}{*}{\makecell[{{l}}]{Emotional\\ Intelligence} }
& \cite{sabour2024emobench} & 2024 & GPT-4, Baichuan, ChatGLM-3 & Proposes EMOBENCH, a comprehensive benchmark for evaluating Emotional Intelligence (EI) of LLMs, emphasizing gaps between LLMs and humans in understanding emotions & Accuracy in emotional understanding tasks & \ding{51} \\ \cmidrule(l){2-7}
& \cite{zhao2024both} & 2024 & Flan-T5, LLaMA-2-Chat & Introduces EIBENCH, a collection of EI-related tasks, and proposes MoEI, a modular EI enhancement method to improve LLMs' EI without compromising their General Intelligence (GI) & Accuracy on EI and GI benchmarks & \ding{55} \\ \cmidrule(l){2-7}
& \cite{wang2023emotional} & 2024 & GPT-4, Claude, LLaMA-based models & Presents SECEU, a psychometric evaluation of Emotional Understanding (EU) for LLMs, showing that GPT-4 achieves higher emotional intelligence scores than 89\% of human participants & Standardized EQ scores, comparison to human performance & \ding{51} \\ \midrule

\multirow{3}{*}{\makecell[{{l}}]{Collaboration} } 
& \cite{liu2023dynamic} & 2023 & GPT-3.5-Turbo, GPT-4 & Introduces DyLAN (Dynamic LLM-Agent Network), a framework that optimizes LLM-agent collaboration through dynamic agent selection, enhancing efficiency and adaptability in complex tasks & Accuracy, \#API & \ding{55} \\ \cmidrule(l){2-7}
& \cite{chen2024llmarena} & 2024 & GPT-4, GPT-3.5, Qwen, LLaMA-2, AgentLM, DeepSeek, SUS, Yi, WizardLM, Vicuna & Presents LLMARENA, a benchmark designed to evaluate the abilities of LLMs in multi-agent, dynamic settings across seven different gaming environments & TrueSkill™, reward & \ding{51} \\ \cmidrule(l){2-7}
& \cite{zhang2024towards} & 2024 & GPT-4, GPT-3.5-Turbo, Qwen, LLaMA-2, AgentLM, and DeepSeek & Proposes ReAd (Reinforced Advantage), a closed-loop feedback mechanism for LLM planners to enhance decision-making and efficiency in multi-agent collaboration & Success rate, environment steps, number of queries & \ding{55} \\ \cmidrule(l){2-7}
& \cite{piatti2024cooperate} & 2024 & GPT-3.5, GPT-4, GPT-4o, Claude-3, LLaMA-2, LLaMA-3, Mistral, Qwen & Introduces GOVSIM, a generative simulation benchmark designed to evaluate the cooperative decision-making abilities of LLM-based agents in multi-agent resource-sharing scenarios & Survival rate and time, total gain, efficiency, Equality, Over-usage & \ding{55} \\ 
\midrule

\multirow{3}{*}{\makecell[{{l}}]{Cultural \\ Competency}}
&    \cite{dudy2024analyzing}      &   2024    &  Mistral, Gemma, LLaMA-2, GPT-3.5-Turbo, GPT-4-Turbo-Preview &  Evaluates how LLMs represent emotions across different cultures, particularly in mixed-emotion scenarios
 &  T-Test &         \ding{51}            \\ \cmidrule(l){2-7}

&       \cite{chang2024benchmarking}      &   2024    &  GPT-4o, Gemini 1.5 Pro, Claude 3.5 Sonnet, LLaMA 3.1, with RAG  &  Presents a benchmark to evaluate LLMs in understanding and processing Hakka cultural knowledge
 &  Accuracy &         \ding{55}            \\ \cmidrule(l){2-7}

 &    \cite{kharchenko2024well}      &   2024    &  GPT-4, GPT-4o, LLaMA 3, Command R+, Gemma  &  Evaluates LLMs ability to understand and reflect cultural values based on Hofstede's Cultural Dimensions framework &  Pearson correlation coefficient, accuracy 
 &         \ding{55}            \\ \cmidrule(l){2-7}

&    \cite{myung2024blend}      &   2024    &  GPT-3.5, GPT-4, Claude-3-Opus, Claude-3-Sonnet, Claude-3-Haiku, PaLM2, Gemini-1.0-Pro, C4AI Command R+, C4AI Command R, Qwen-1.5, SeaLLM, HyperCLOVA-X, Aya-23, Aya-101  &  Designs BLEND benchmark to evaluate LLMs on their cultural knowledge of everyday life across diverse regions and languages
 &  Cultural knowledge, short-answer questions, multiple-choice questions, performance gap, local languages, English performance, cultural sensitivity &         \ding{51}         \\

\bottomrule
\label{table:paper_summary2}
\end{tabular}
}
\end{table}

\begin{table}[ht]
\label{tab:domains_indiv}
\caption{Human-centric LLM studies in the individual domains of behavioral science, psychology, and linguistics.}
\resizebox{\textwidth}{!}{%
\begin{tabular}{@{}cccccccc@{}}
\toprule
\textbf{Field}              & \textbf{Paper} & \textbf{Year} & \textbf{LLM(s) Used}                                                              & \textbf{Method}                                              & \textbf{Problem Addressed}                                                                                                                   & \textbf{Human Phenomenon/Theory}                                                                                 & \textbf{Research Goal} \\ \midrule

\multirow{11}{*}{\begin{tabular}[c]{@{}c@{}}Behavioral\\ Science\end{tabular}}
                          & \cite{he2024ai}                                & 2024          & LLaMA-2                                                                  & \begin{tabular}[c]{@{}c@{}}Multi-Agent\\ Prompting\\ Human-in-the-Loop\end{tabular}                        & \begin{tabular}[c]{@{}c@{}}Incorporates LLM into a \\ collaborative work scenario\end{tabular}                                                     & Collaborative ideation & Capability Discovery                          \\ \cmidrule(l){2-8}
                          & \cite{shaer2024aiaugmented}                                    & 2024          & GPT-3/4                                                                   & \begin{tabular}[c]{@{}c@{}}Multi-Agent\\ Prompting\\ Human-in-the-Loop\end{tabular}                        & \begin{tabular}[c]{@{}c@{}}Explores how LLMs can assist \\ with collaborative writing\end{tabular}                                                      & Brainstorming theories & Capability Discovery           \\ \cmidrule(l){2-8} 
                          & \cite{guo2024embodied}                                         & 2024          & GPT-4/3.5-turbo                                                        & \begin{tabular}[c]{@{}c@{}}Multi-Agent\\ Prompting\\ Human-in-the-Loop\end{tabular}                        & \begin{tabular}[c]{@{}c@{}}Improves efficiency of \\ organizational structure\end{tabular}                     & Organization theory & Capability Discovery                             \\ \cmidrule(l){2-8} 
                          & \cite{liu2024peergpt}    & 2024          & GPT-3.5                & \begin{tabular}[c]{@{}c@{}}Multi-Agent\\ Prompting\\ Human-in-the-Loop\end{tabular}                        & \begin{tabular}[c]{@{}c@{}}Determines the role of LLM in \\ collaborative learning environments\end{tabular}                           & Collaborative Learning & Capability Discovery                          \\ \cmidrule(l){2-8} 
                          & \cite{park2023generative}                  & 2023          & GPT-3.5-Turbo                                                                  & \begin{tabular}[c]{@{}c@{}}Multi-Agent\\ Prompting\end{tabular}                        & \begin{tabular}[c]{@{}c@{}}Creates a simulated village \\ of LLM agents\end{tabular}                                                              & Cognitive architectures & Simulation             \\ \cmidrule(l){2-8} 
                          & \cite{li2023camel}                                    & 2023          & \begin{tabular}[c]{@{}c@{}}GPT-4/3.5-turbo,\\ LLaMA\end{tabular}                                                    & \begin{tabular}[c]{@{}c@{}}Multi-agent\\Prompting\\Fine-Tuning, \\ Human-in-the-Loop\end{tabular}                      & \begin{tabular}[c]{@{}c@{}}Facilitates autonomous cooperation\\ among communicative agents\end{tabular}                                           & Cooperation   & Capability Discovery   \\ \cmidrule(l){2-8} 
                          & \cite{shaikh2023rehearsal}                          & 2023          & GPT-4                                                                          & \begin{tabular}[c]{@{}c@{}}Multi-Agent\\ Prompting\end{tabular}                        & \begin{tabular}[c]{@{}c@{}}Evaluates LLM's ability to \\ perform conflict resolution\end{tabular}                                            & Interest, rights, power (IRP) & Simulation                      \\ \cmidrule(l){2-8} 
                          & \cite{park2022social}                                      & 2022          & GPT-3                                                                          & \begin{tabular}[c]{@{}c@{}}Multi-Agent\\ Prompting\end{tabular}                        & \begin{tabular}[c]{@{}c@{}}Evaluates LLM's ability to \\ simulate social media dynamics\end{tabular}                                    & Community norms & Simulation                                    \\ \cmidrule(l){2-8} 
                          & \cite{wang2024unleashing} & 2024          & \begin{tabular}[c]{@{}c@{}}GPT-4/3.5-Turbo,\\ Llama2-13b\end{tabular}                                        & \begin{tabular}[c]{@{}c@{}}Multi-Agent\\ Prompting\end{tabular}                        & \begin{tabular}[c]{@{}c@{}}Evaluates LLM's ability to \\ collaboratively solve problems\end{tabular}                                  & Cognitive Synergy & Simulation    \\ \cmidrule(l){2-8}                             
                         &   \cite{gao2023s3}                                & 2023          & \begin{tabular}[c]{@{}c@{}}GPT-based LLMs\end{tabular}                                                                  & \begin{tabular}[c]{@{}c@{}}Multi-Agent\\ Prompting\end{tabular}                        & \begin{tabular}[c]{@{}c@{}}Simulates social\\ network dynamics\end{tabular}                                                     & \begin{tabular}[c]{@{}c@{}}Social network\\ theory\end{tabular} & Simulation           
                          
                          \\ \midrule
\multirow{7}{*}{Psychology}  &   \cite{strachan2024testing}             & 2024          & \begin{tabular}[c]{@{}c@{}}GPT-4\\ LLaMA-2\end{tabular}                        & \begin{tabular}[c]{@{}c@{}}Prompting\\ Human-in-the-Loop\end{tabular}                                                       & Tests theory of mind in LLMs                                                                                                             & Theory of mind                                                                                         & Capability Discovery        \\ \cmidrule(l){2-8} 
                             &   \cite{stella2023using}             & 2023          & GPT-3                                                                          & Prompting                                                       & \begin{tabular}[c]{@{}c@{}}Distinguishes intrinsic knowledge elaboration\\ from knowledge copy-pasting in LLMs\end{tabular}               & Human bias                                                                                             & Capability Discovery        \\ \cmidrule(l){2-8} 
                             &   \cite{chuang2024wisdom}            & 2024          & ChatGPT                                                                        & \begin{tabular}[c]{@{}c@{}}Multi-Agent\\ Prompting\\ Fine-tuning\\ Human-in-the-Loop\end{tabular}  & Examines collective intelligence in LLM-based agents                                                                                      & Collective intelligence                                                                                & Simulation \\ \cmidrule(l){2-8} 
                             &   \cite{dillion2023can}            & 2023          & GPT-3.5                                                                        & \begin{tabular}[c]{@{}c@{}}Prompting\\ Human-in-the-Loop\end{tabular}                                                       & \begin{tabular}[c]{@{}c@{}}Explores how LLMs could become a substitute\\ for the psychology study participants\end{tabular}                  & Game theory                                                                                            & Capability Discovery \\ \cmidrule(l){2-8} 
                             &   \cite{peters2023large}             & 2024          & GPT-4/3.5                                                                      & \begin{tabular}[c]{@{}c@{}}Prompting\\ Human-in-the-Loop\end{tabular}                                                       & \begin{tabular}[c]{@{}c@{}}Examines whether LLMs can accurately infer the\\ psychological dispositions of social media users\end{tabular} & Big five personality traits theory                                                                     & Capability Discovery        \\ \cmidrule(l){2-8} 
                             &   \cite{zhang2023exploring}             & 2024          & ChatGPT                                                                        & \begin{tabular}[c]{@{}c@{}}Multi-Agent\\ Prompting\\ Human-in-the-Loop\end{tabular} & Explores collaboration mechanisms for LLM Agents                                                                                          & Collective intelligence                                                                                & Simulation        \\ \cmidrule(l){2-8} 
                             &   \cite{Demszky2023}             & 2023          & GPT-3                                                                          & \begin{tabular}[c]{@{}c@{}}Prompting\\ Fine-tuning\\ Human-in-the-Loop\end{tabular}  & \begin{tabular}[c]{@{}c@{}}Explores how LLMs can be integrated\\ into psychological research and practice\end{tabular}                    & AI bias                                                                                                & Capability Discovery \\ \midrule
\multirow{3}{*}{Linguistics} &   \cite{mahowald2024dissociating}             & 2024          & GPT-4/3                                                                        & Prompting                                                       & \begin{tabular}[c]{@{}c@{}}Evaluating LLMs using a distinction between\\ formal and functional linguistic competence\end{tabular}          & Social Reasoning                                                                                         & Capability Discovery        \\ \cmidrule(l){2-8} 
                             &   \cite{lai2023chatgpt}             & 2023          & GPT-3.5                                                                        & Prompting                                                       & \begin{tabular}[c]{@{}c@{}}Evaluates ChatGPT's performance\\ in multilingual NLP tasks\end{tabular}                                       & Misinformation and plagiarism                                                                          & Capability Discovery        \\ \cmidrule(l){2-8} 
                             &   \cite{pavlick2023symbols}             & 2023          & GPT-3                                                                          & Prompting                                                       & \begin{tabular}[c]{@{}c@{}}Evaluates the potential of LLMs to serve as\\ models of language understanding in humans\end{tabular}          & Cognitive science                                                                                      & Capability Discovery        \\ \bottomrule
        \label{tbl:econ_socio_psych_ling}
\end{tabular}
}
\end{table}

\begin{table}[ht]
\label{tab:domains_collective}
\caption{Human-centric LLM studies in the collective domains of political science, economics, and sociology.}
\resizebox{\textwidth}{!}{%
\begin{tabular}{@{}cccccccc@{}}
\toprule
\textbf{Field}              & \textbf{Paper} & \textbf{Year} & \textbf{LLM(s) Used}                                                              & \textbf{Method}                                              & \textbf{Problem Addressed}                                                                                                                     & \textbf{Human Phenomenon/Theory}                                                                                 & \textbf{Research Goal} \\ \midrule

\multirow{11}{*}{\begin{tabular}[c]{@{}c@{}}Political\\ Science\end{tabular}}
                          & \cite{hua2024war}                              & 2024          & \begin{tabular}[c]{@{}c@{}}GPT-4/3.5-turbo,\\ Claude-2\end{tabular}                               & \begin{tabular}[c]{@{}c@{}}Multi-Agent\\ Prompting \\ Fine-Tuning\end{tabular}           & \begin{tabular}[c]{@{}c@{}}Explores where LLM system \\ simulations can replicate \\ historical events\end{tabular}       & IR Theory, Historical Determinism & Simulation                  \\ \cmidrule(l){2-8}
                          & \cite{moghimifar2024modelling}                                                         & 2024          & \begin{tabular}[c]{@{}c@{}}GPT-3.5-turbo,\\ LLaMa(7b \& 13b)\end{tabular}                               & \begin{tabular}[c]{@{}c@{}}Multi-Agent\\ Prompting\end{tabular}   & \begin{tabular}[c]{@{}c@{}}Predicts the formation of \\ political coalitions\end{tabular}                           & N/A & Simulation                                                \\ \cmidrule(l){2-8} 
                          & \cite{taubenfeld2024systematic}                                          & 2024          & \begin{tabular}[c]{@{}c@{}}Mistral, Solar,\\ Instruct-GPT\end{tabular}                                           & \begin{tabular}[c]{@{}c@{}}Multi-Agent\\ Prompting \\ Fine-Tuning\end{tabular}           & \begin{tabular}[c]{@{}c@{}}Simulates debates to \\ understand biases in LLMs\end{tabular}         & Echo chamber theory & Simulation                                \\ \cmidrule(l){2-8} 
                          & \cite{hackenburg2024evaluating}            & 2024          & GPT-4                                                                          & Prompting               & \begin{tabular}[c]{@{}c@{}}Evaluates how well LLM can create \\ personalized political messaging\end{tabular}             & Persuasion & Content Generation                                 \\ \cmidrule(l){2-8}  
                          & \cite{goldstein2024how}                                              & 2024          & GPT-3                                      & Prompting          & \begin{tabular}[c]{@{}c@{}}Evaluates how well LLM can create \\ persuasive political messaging\end{tabular}        & Persuasion & Content Generation                                 \\ \cmidrule(l){2-8}  
                          & \cite{zhu2023cana}               & 2023          & GPT-3.5-turbo                                                                  & Prompting                                     & \begin{tabular}[c]{@{}c@{}}Evaluates accuracy of LLMs \\ on political labeling tasks\end{tabular}                        & N/A & Labeling                                                  \\ \cmidrule(l){2-8} 
                          & \cite{urman2023silence}  & 2023     & \begin{tabular}[c]{@{}c@{}}ChatGPT,\\ Bing Chat, Bard\end{tabular}                                                       & Prompting             & \begin{tabular}[c]{@{}c@{}}Assess LLM's tendency to \\ produce misinformation\end{tabular}              & N/A & Labeling                                                  \\ \cmidrule(l){2-8} 
                          & \cite{lan2024stance}                            & 2023          & GPT-3.5-turbo                                                                  & \begin{tabular}[c]{@{}c@{}}Prompting\end{tabular}                            & \begin{tabular}[c]{@{}c@{}}Detects political standpoints \\ using LLM\end{tabular}             & N/A & Labeling                                                  \\ \cmidrule(l){2-8} 
                          & \cite{zhang2024llmdriven}            & 2024          & BART                                                                           & Pre-training                                   & \begin{tabular}[c]{@{}c@{}}Detects political standpoints \\ using LLM\end{tabular}                   & N/A & Labeling                                                  \\ \cmidrule(l){2-8}  
                          & \cite{li2024mitigating}                   & 2024          & \begin{tabular}[c]{@{}c@{}}GPT-3.5-turbo,\\ LLaMA2\end{tabular}                                                      & \begin{tabular}[c]{@{}c@{}}Prompting\\ Calibration\end{tabular}                        & \begin{tabular}[c]{@{}c@{}}Detects political standpoints and \\ mitigates biases using LLM\end{tabular}                     & N/A & Debiasing                                                  \\ 
                        \\ \midrule
                        
                        \multirow{10}{*}{Economics} 
                          & \cite{guo2023gpt}          & 2023          & GPT-4                                                                          & Prompting              & 
                          
                          \begin{tabular}[c]{@{}c@{}}Investigates LLM in \\ strategic game experiments\end{tabular} & \begin{tabular}[c]{@{}c@{}}Game theory\\ Prisoner’s dilemma\end{tabular} & Capability Discovery                   \\ \cmidrule(l){2-8}
                          & \cite{akata2023playing}                                 & 2023          & \begin{tabular}[c]{@{}c@{}}GPT-3/3.5/4\end{tabular}                                                          & Prompting                                     & \begin{tabular}[c]{@{}c@{}}Studies the cooperation and \\ coordination behavior of LLM\end{tabular}         & \begin{tabular}[c]{@{}c@{}}Game theory\\ Prisoner’s dilemma\end{tabular} & Capability Discovery                   \\ \cmidrule(l){2-8} 
                          & \cite{weiss2023rethinking}                                    & 2023          & \begin{tabular}[c]{@{}c@{}}GPT-3.5/4,\\ LLaMA-2\end{tabular}                                    & \begin{tabular}[c]{@{}c@{}}Multi-Agent\\ Prompting\end{tabular}                        & \begin{tabular}[c]{@{}c@{}}Simulates game scenarios \\ in a digital marketplace\end{tabular} &Game Theory  & Simulation                     \\ \cmidrule(l){2-8} 
                          & \cite{zhao2024competeai}          & 2024          & GPT-4                                                                          & \begin{tabular}[c]{@{}c@{}}Multi-Agent\\ Prompting\end{tabular}                        & \begin{tabular}[c]{@{}c@{}}Simulates a virtual town \\ for market competition\end{tabular}                                   & \begin{tabular}[c]{@{}c@{}}Market competition\\ Social learning theory\\ Consumer behavior\end{tabular} & Capability Discovery   \\ \cmidrule(l){2-8} 
                          & \cite{horton2023large}                         & 2023          & GPT-3                                                                          & Prompting                                     & \begin{tabular}[c]{@{}c@{}}Replicates classic \\ social science studies\end{tabular}                    & \begin{tabular}[c]{@{}c@{}}Game theory\\ Social preferences\end{tabular} & Capability Discovery                   \\ \cmidrule(l){2-8} 
                          & \cite{shang2024synergyofthoughts}                        & 2024          & \begin{tabular}[c]{@{}c@{}}GPT-3.5/4, Mistral \\ LLaMA, Yi \\ PaLM2, Gemini1pro\end{tabular}               & \begin{tabular}[c]{@{}c@{}}Multi-Agent\\ Multi-Model \\ Framework\end{tabular}            & \begin{tabular}[c]{@{}c@{}}Increases efficiency \\ of LLM deployment\end{tabular}  & Human Cognition & Efficiency             \\ \cmidrule(l){2-8} 
                          & \cite{li2024econagent}           & 2024          & GPT-3.5-turbo                                                                  & \begin{tabular}[c]{@{}c@{}}Multi-Agent\\ Prompting\end{tabular}                        & \begin{tabular}[c]{@{}c@{}}Simulates decision-making \\ mechanisms for macroeconomic trends\end{tabular}    & Macroeconomic theory & Simulation                              \\ \cmidrule(l){2-8} 
                          & \cite{suzuki2024evolutionary}                  & 2024          & GPT-3.5                                                                        & Prompting                                     & \begin{tabular}[c]{@{}c@{}}Simulates the evolution of \\ human collaborative behavior\end{tabular}  & \begin{tabular}[c]{@{}c@{}}Game theory\\ Prisoner’s dilemma\end{tabular} & Simulation  \\ \midrule
\multirow{8}{*}{Sociology}   &  \cite{liu2023multilingual}              & 2024          & LLaMA-2                                                                        & Prompting                                                       & \begin{tabular}[c]{@{}c@{}}Incorporates cultural reasoning\\ into multilingual LLMs\end{tabular}                                          & Cultural common ground                                                                                 & Debiasing         \\ \cmidrule(l){2-8} 
                             &  \cite{kotek2023gender}              & 2023          & GPT-3.5                                                                        & Prompting                                                       & Explores gender bias in LLMs                                                                                                              & \begin{tabular}[c]{@{}c@{}}Gender bias\\ Stereotyping\end{tabular}                                     & Quantifying Bias           \\ \cmidrule(l){2-8} 
                             &   \cite{tao2023auditing}             & 2024          & GPT-3/3.5/4/4o                                                                 & \begin{tabular}[c]{@{}c@{}}Prompting / \\ Human-in-the-loop \end{tabular}                                                    & Explores cultural bias in LLMs                                                                                                            & Inglehart-Welzel cultural map                                                                          & Quantifying Bias     \\ \cmidrule(l){2-8} 
                             % &   \cite{duan2023denevil}             & 2024          & LLaMA                                                                          & \begin{tabular}[c]{@{}c@{}}Prompting\\ Finetuning\end{tabular}  & Aligns the ethical values of LLMs                                                                                                        & Moral foundations theory                                                                               & Human in the Loop \\ \cmidrule(l){2-8} 
                             &   \cite{scherrer2024evaluating}             & 2023          & \begin{tabular}[c]{@{}c@{}}GPT-4\\ Claude\\ PaLM-2\end{tabular}                & Prompting                                                       & Evaluates the moral beliefs encoded in LLMs                                                                                               & Common morality framework                                                                              & Capability Discovery        \\ \cmidrule(l){2-8} 
                             &   \cite{rao2024normad}             & 2024          & \begin{tabular}[c]{@{}c@{}}GPT-4/3.5-turbo\\ LLaMA-2\\ Mistral-7B\end{tabular} & Prompting                                                       & Evaluates the cultural adaptability of LLMs                                                                                               & \begin{tabular}[c]{@{}c@{}}Kohlberg’s theory of morality\\ Hofstede’s cultural dimensions\end{tabular} & Capability Discovery        \\ \cmidrule(l){2-8} 
                             &   \cite{dwivedi2023eticor}             & 2023          & \begin{tabular}[c]{@{}c@{}}GPT-3.5\\ Falcon40B\end{tabular}                    & Prompting                                                       & \begin{tabular}[c]{@{}c@{}}Evaluates LLM's sensitivity to\\ region-specific etiquettes\end{tabular}                                    & Ethical norms                                                                                          & Capability Discovery        \\ \cmidrule(l){2-8} 
                             &   \cite{zheng2024ali}             & 2023          & \begin{tabular}[c]{@{}c@{}}GPT-4/3.5\\ LLaMA-2\end{tabular}                    & \begin{tabular}[c]{@{}c@{}}Multi-Agent\\ Prompting\end{tabular} & \begin{tabular}[c]{@{}c@{}}Evaluates and refining the\\ alignment of LLMs with human values\end{tabular}                                  & \begin{tabular}[c]{@{}c@{}}Human morality\\ Stereotyping\end{tabular}                                  & Capability Discovery

 \\ \bottomrule
 \label{tbl:behavioralpolitical}
\end{tabular}%
}
\end{table}

\end{document}